\documentclass[runningheads]{llncs}

\usepackage{algorithm2e}
\SetKwInput{KwInput}{Input}     
\SetKwInput{KwOutput}{Output} 
\usepackage{amsmath}
\usepackage{amssymb}
\usepackage{multirow}
\usepackage{graphicx}
\usepackage{booktabs}

\usepackage{soul}
\usepackage{subcaption}
\usepackage{theorem}
\author{Pierre Nodet\inst{1,2} \and Vincent Lemaire \inst{1} \and Alexis Bondu\inst{1} \and Antoine Cornu\'ejols\inst{2}}

\authorrunning{P. Nodet et al.}
\institute{Orange Innovation, France \and AgroParisTech, France }

\begin{document}

\title{Biquality Learning: a Framework to Design Algorithms Dealing with Closed-Set Distribution Shifts}

%\title[Design of Algorithms Dealing with Closed-Set Distribution Shifts]{Biquality Learning: a Framework to Design Algorithms Dealing with Closed-Set Distribution Shifts}

\maketitle

\begin{abstract}
Training machine learning models from data with weak supervision and dataset shifts is still challenging.
Designing algorithms when these two situations arise has not been explored much, and existing algorithms cannot always handle the most complex distributional shifts.
We think the biquality data setup is a suitable framework for designing such algorithms. Biquality Learning assumes that two datasets are available at training time:
a trusted dataset sampled from the distribution of interest and the untrusted dataset with dataset shifts and weaknesses of supervision (aka distribution shifts).
The trusted and untrusted datasets available at training time make designing algorithms dealing with any distribution shifts possible.
We propose two methods, one inspired by the label noise literature and another by the covariate shift literature for biquality learning.
We experiment with two novel methods to synthetically introduce concept drift and class-conditional shifts in real-world datasets across many of them.
We opened some discussions and assessed that developing biquality learning algorithms robust to distributional changes remains an interesting problem for future research.
\end{abstract}




%-------------------------------------------------------
\section{Introduction}
\label{introduction}
%-------------------------------------------------------

Supervised machine learning has been studied and explored extensively during the last decades, and both theoretical and experimental solutions exist to accomplish this task \cite{hastie2009elements}. Weakly supervised machine learning (WSL) has not reached this state yet. WSL is the machine learning field where algorithms learn a model from data with weak supervision instead of strong supervision. Multiple weak supervisions have been identified in \cite{zhou:2017} such as \textit{inaccurate supervision} when samples are mislabeled, \textit{inexact supervision} when labels are not adapted to the classification task, or \textit{incomplete supervision} when labels are missing which reflects the inadequacy of the available labels in the real world. For every kind of weak supervision, assumptions are needed to design sound algorithms, especially on the corruption model, the generative process behind the weakness of supervision. 

Weaknesses in supervision are one facet of weakly supervised learning, and dataset shifts are another. Dataset shifts happen when the data distribution observed at training time is different from what is expected from the data distribution at testing time \cite{moreno2012unifying}. This distribution change can take multiple forms, such as a change in the distribution of a single feature, a combination of features, or the concept to be learned. Thus, the common assumption that the training and testing data follow the same distributions is often violated in real-world applications. Again, designing algorithms to handle dataset shifts usually requires assumptions on the nature of the shift \cite{david2010impossibility}.

We believe that the biquality data setup proposed in \cite{nodet2020importance} is a suitable framework to design algorithms capable of handling both dataset shifts and weaknesses of supervision simultaneously.

Biquality Learning assumes that two datasets are available at training time: a \textit{trusted dataset} $D_T$ and an \textit{untrusted dataset} $D_U$ both composed of labeled samples $(x_i, y_i) \in \mathcal{X}\times\mathcal{Y}$. These datasets share the same features $\mathcal{X}_T=\mathcal{X}_U$ and the same set of labels $\mathcal{Y}_T=\mathcal{Y}_U$, having a closed-set of features $\mathcal{X}$ and labels $\mathcal{Y}$ with $K=\vert\mathcal{Y}\vert$ classes. However, the two datasets differ in terms of the joint distribution $\mathbb{P}_T(X,Y)\neq\mathbb{P}_U(X,Y)$ where the trusted distribution is the distribution of interest.
% $(x_i, y_i)_{i \in [\![1, \vert D \vert ]\!]} \in \mathcal{X}\times\mathcal{Y}$

In practice, it is often the case that the trusted dataset is not large enough to learn an efficient model to estimate $\mathbb{P}_T(X,Y)$. By contrast, it is generally easy to get a large enough untrusted dataset that enables to correctly estimate $\mathbb{P}_U(X,Y)$. However, this data distribution could be completely different from the one of interest, $\mathbb{P}_T(X,Y)$. In the biquality data setup, there is no assumption on the difference in joint distribution between the two datasets, and it can cover a wide range of known problems. From the Bayes Formula:
\begin{equation}
\label{distribution-shift-equation}
    \mathbb{P}(X,Y) = \mathbb{P}(X \mid Y)\mathbb{P}(Y) = \mathbb{P}(Y \mid X)\mathbb{P}(X)
\end{equation}
distribution shift covers covariate shift $\mathbb{P}_T(X)\neq\mathbb{P}_U(X)$, concept drift $\mathbb{P}_T(Y \mid X)\neq\mathbb{P}_U(Y \mid X)$, class-conditional shift $\mathbb{P}_T(X \mid Y)\neq\mathbb{P}_U(X \mid Y)$ and prior shift $\mathbb{P}_T(Y)\neq\mathbb{P}_U(Y)$.\\

The Biquality Data setup typically occurs in three scenarios in practice.
\begin{enumerate}
    \item The first scenario corresponds to the case where annotating samples is expansive to the point of being prohibitive to label an entire dataset, but labeling a small part of the dataset is doable. It is typically the case in Fraud Detection and in Cyber Security where labeling samples require complex forensics from domain experts. There, the rest of the dataset is usually labeled by hand-engineered rules which might not perfectly fit the classification task and the labels can not properly be trusted \cite{ratner2020snorkel}.
    \item The second scenario happens when there are data shifts during the labeling process over the course of time. For example in MLOps \cite{kreuzberger2022machine}, when a model is first learned on clean data and then deployed in production, predictions can be used to learn an updated model \cite{7502263}; these predictions may be faulty and need to be dealt with. A second example in MLOps occurs when newly clean data is acquired to retrain a deployed model; the most recent clean data is considered trusted and the old clean data is considered untrusted when dataset shifts occurred \cite{gama2014survey}.
    \item The third scenario occurs when multiple annotators are responsible for dataset labeling. In natural language processing (NLP) products, for example, multiple annotators follow labeling guidelines to annotate verbatims. The efficiency of annotators to follow these guidelines may vary and might not be trusted. However, if one annotator can be trusted, all the other annotators can be set as untrusted and the biquality data setup may apply. Especially, considering each untrusted annotator against the trusted annotator can be viewed as a biquality learning task \cite{yuen2011survey}.
\end{enumerate}

Having the trusted and untrusted datasets available at training time makes it possible to design algorithms dealing with closed-set distribution shifts. If algorithms designed for biquality data with concept drift only have been explored recently \cite{nodet2021weakly}, algorithms that deal with distribution shift are still behind. In this paper, we propose two biquality approaches that adapt methods from either the covariate shift literature or the concept drift literature in order to deal with both corruptions simultaneously. 

Multiple biquality learning algorithm designs have been identified in \cite{nodet2020importance}. They have been divided into three main families based on how they modify instances to correct the global learning procedure. Untrusted instances can be (i) relabeled correctly, (ii) modified in the feature space, or (iii) reweighted such that the untrusted dataset seems sampled from the trusted distribution $\mathbb{P}_T(X,Y)$. We propose here two corresponding algorithms for the third case of importance reweighting.
%as instance reweighting is supported by a majority of classifier families \cite{quinlan1996learning, 10.1145/130385.130401, rosenblatt1961principles}.
% \textcolor{green}{(argument étrange, puisque les deux autres façon de faire sont également suportées par tous les classifieurs (corriger X ou y))}
% and if not, importance resampling can be used \cite{an2020resampling, seiffert2008resampling}.

In Section \ref{related_work}, a brief state-of-the-art relates what has already been achieved in biquality learning. Then, Section \ref{reweighting} further focuses on the state-of-the-art of importance reweighting on biquality learning. Section \ref{irbl2} and Section \ref{kpdr} introduce our proposals to use classifiers to reweight untrusted instances for biquality learning with distribution shifts. Then, Section \ref{experiments} describes the experiments that evaluate the efficiency of our proposed approaches on real datasets and corruptions. Section \ref{results} presents the results of the proposed experiments. Finally, Section \ref{discussions} open some discussions about the presented results before concluding.

\section{Related work}
\label{related_work}

Machine learning algorithms on biquality data have been developed in many different sub-domains of weakly supervised learning. Some of these sub-domains are robust learning to label noise, learning under covariate shift, or transfer learning. Because these subdomains expect different corruptions, not all algorithms designed for some subcases of biquality learning will work in a more general setting with distribution shifts.

For example, Gold Loss Correction (GLC) \cite{Hendrycks2018} and Importance Reweighting for Biquality Learning (IRBL) \cite{nodet2020importance} are algorithms designed to specifically deal with a concept drift between the trusted and the untrusted datasets. GLC, on the one hand, corrects the learning procedure on the untrusted dataset using a noise transition matrix between the trusted and untrusted concept. IRBL, on the other hand, reweight untrusted instances using an estimation of the ratio of both concepts. These algorithms are not theoretically designed to handle covariate shift; nevertheless, they could be empirically efficient on this task and serve as a reference.

Another group of algorithms, such as Kernel Mean Matching (KMM) \cite{gretton2009covariate}, and Probabilistic Density Ratio Estimation (PDR) \cite{bickel2007discriminative}, only deals with covariate shift between the two datasets. These algorithms seek to reweight untrusted instances such that the distribution of features between the two datasets is equivalent. KMM minimizes the difference of the features mean in a reproducing kernel Hilbert space. PDR learns the classification task of predicting if an instance is untrusted or not and uses the predicted probability of being an untrusted instance as the weight. These algorithms are not designed to handle concept drift and will serve as references too.

However, a recent proposal aims to adapt these algorithms to the biquality framework with distribution shift \cite{fang2020rethinking}. They proposed to use one density ratio estimation algorithm per class, which, when combined, corrects the distribution shift. They also proposed to transform the joint density ratio estimation problem by combining the features and labels of the data into a new feature space that allows for a single density ratio estimation algorithm. These adapted algorithms will also serve as competitors.

Finally, recent approaches such as Learning to Reweight (L2RW) \cite{ren2018learning}, or Meta-Weight-Net (MWNet) \cite{shu2019mwnet} based on deep learning and meta-learning have not been tested in this paper. Indeed, they require a class of algorithms with a differentiable and incremental learning procedure that does not fit most popular families of classifiers, such as gradient boosting trees. They are left to be tested in future works.

Biquality Data is not a new setup per see, as previous work exists on this setup going back to \cite{jiang2018mentornet} to the best of our knowledge. Each previous work was carried out in different sub-domains of weakly supervised learning and thus achieved different goals based on different setups \cite{Hendrycks2018,shu2019mwnet,ren2018learning,zheng2021mlc,jiang2018mentornet}. These setups used different terms, definitions, hypotheses, and requirements but still sought to solve the same fundamental problem of biquality learning. Only recently, some efforts have been done to provide clear and concise definitions of the biquality learning framework \cite{nodet2020importance}. We propose in this paper to extend it to include dataset shifts. This extension is, to the best of our knowledge, a new problem to tackle that few tried already \cite{fang2020rethinking}, limiting existing prior literature.

\section{Reweighting for distribution shift}
\label{reweighting}

The previous Section introduced the most common algorithms used in machine learning for biquality data. Most of them are based on instance reweighting, and specifically, on estimating the Radon-Nikodym Derivative (RND) \cite{Nikodym1930} of $\mathbb{P}_T(X,Y)$ with respect to $\mathbb{P}_U(X,Y)$.

\begin{theorem}[Radon-Nikdoym-Lebesgue theorem \cite{rudin1975analyse}]
\label{rnlt}
Let $\mu$ and $\nu$ two positive $\sigma$-finite measures defined on a measurable space $(X, \mathcal{A})$ with $\nu$ being absolutely continuous with respect to $\mu$.
Then there exists a unique positive measurable function $f$ defined on $X$ such that:
\begin{equation}
\forall A \in \mathcal{A}, \nu(A) = \int_A f d\mu
\end{equation}
\end{theorem}

Typically, machine learning datasets form measurable spaces, and probability densities are positive finite measures on these measurable spaces. Assuming that $\mathbb{P}_T(X,Y)$ is absolutely continuous with respect to $\mathbb{P}_U(X,Y)$, then the RND exists, is unique and equals to $\frac{\text{d}\mathbb{P}_T(X,Y)}{\text{d}\mathbb{P}_U(X,Y)}$.

Equation \ref{risk-loss} shows that minimizing the reweighted empirical risk by the RND on the untrusted data is equivalent to minimizing the empirical risk on trusted data.

\begin{equation}
\begin{aligned}
&R_{(X,Y)\sim T,L}(f) = \mathbb{E}_{(X,Y)\sim T}[L(f(X),Y)]\\
&\qquad = \int L(f(X),Y)\,\text{d}\mathbb{P}_T(X,Y)\\
&\qquad = \int \frac{\text{d}\mathbb{P}_T(X,Y)}{\text{d}\mathbb{P}_U(X,Y)}L(f(X),Y)\,\text{d}\mathbb{P}_U(X,Y)\\
&\qquad = \mathbb{E}_{(X,Y)\sim U}[\frac{\mathbb{P}_T(X,Y)}{\mathbb{P}_U(X,Y)}L(f(X),Y)]\\
&\qquad = \mathbb{E}_{(X,Y)\sim U}[\beta L(f(X),Y)]\\
&\qquad = R_{(X,Y)\sim U,\beta L}(f)\\
\end{aligned}
\label{risk-loss}
\end{equation}

However, estimating the RND can be a difficult task, especially in the case of distribution shift where the joint distribution ratio $\beta$ needs to be estimated. Proposals have been made to ease this estimation.

A \textit{first proposal} has been made in IRBL \cite{nodet2020importance} which focused first on the concept drift between datasets using the Bayes Formula:

\begin{equation}
\label{irbl-equation}
    \beta(X,Y) =
    \frac{\mathbb{P}_T(X,Y)}{\mathbb{P}_U(X,Y)} = \frac{\mathbb{P}_T(Y \mid X)\mathbb{P}_T(X)}{\mathbb{P}_U(Y \mid X)\mathbb{P}_U(X)}
\end{equation}

Their proposed algorithm is based on the decomposition of the joint density ratio estimation task into three sub-tasks. The first one is to estimate the trusted concept $\mathbb{P}_T(Y \mid X)$, and is done by learning a classifier on the trusted dataset. The second task is to estimate the untrusted concept $\mathbb{P}_U(Y \mid X)$, which is done by learning a classifier on the untrusted dataset. And the third task about density ratio estimation $\frac{\mathbb{P}_T(X)}{\mathbb{P}_U(X)}$ was skipped as no covariate shift was introduced in their benchmark, but it is a well known and solved machine learning task \cite{sugiyama2012density}.

A \textit{second proposal} has been made in \cite{fang2020rethinking} which focused on the covariate shift between datasets using the Bayes Formula differently:

\begin{equation}
\label{kdr-equation}
    \beta(X,Y) =
    \frac{\mathbb{P}_T(X,Y)}{\mathbb{P}_U(X,Y)} = \frac{\mathbb{P}_T(X \mid Y)\mathbb{P}_T(Y)}{\mathbb{P}_U(X \mid Y)\mathbb{P}_U(Y)}
\end{equation}

In their proposed algorithm, the joint density ratio estimation task has been decomposed into $K$-tasks where $K$ is the number of classes to predict. For each class, only samples of the given class are selected on both datasets, such that the samples are drawn from the $\mathbb{P}(X \mid Y)$ distribution. Then, a density ratio estimation procedure usually employed to estimate $\frac{\mathbb{P}_T(X)}{\mathbb{P}_U(X)}$ is learned on these sub-datasets to estimate $\frac{\mathbb{P}_T(X \mid Y)}{\mathbb{P}_U(X \mid Y)}$, effectively handling distribution shift from Equation \ref{kdr-equation}. As it uses $K$ density ratio algorithms, this generic approach will be named $K$-DensityRatio ($K$-DR) in the rest of the paper.

Finally, a \textit{last approach} is to focus on the density ratio estimation task by finding a deterministic and invertible transformation $f$ as proposed in \cite{fang2020rethinking}:
\begin{equation}
    \beta(X,Y) =
    \frac{\mathbb{P}_T(X,Y)}{\mathbb{P}_U(X,Y)} =
    \frac{\mathbb{P}_T(Z)}{\mathbb{P}_U(Z)},\quad Z = f(X,Y)
\end{equation}
An example of such transformation \cite{fang2020rethinking} is the classification loss of a model learned on the biquality data. One density ratio estimation procedure is done on these new features $\mathcal{Z}$ to directly estimate $\frac{\mathbb{P}_T(Z)}{\mathbb{P}_U(Z)}$.

IRBL has experimentally proved to efficiently solve the biquality learning task on tabular data \cite{nodet2020importance}. However, the experiments were conducted on corruptions only affecting the untrusted concept $\mathbb{P}(Y \mid X)$ and not the joint distribution $\mathbb{P}(X,Y)$. We propose here to adapt IRBL to handle distribution shifts by solving the third task of density ratio estimation with a probabilistic classifier. 
Moreover, we propose a new version of $K$-DR using probabilistic classifiers to solve the $K$ density ratio estimation tasks. This proposition is driven by the desire to reuse efficient tricks from IRBL and to rely on a non parametric approach by contrast to the original proposal \cite{fang2020rethinking}.

\section{First proposed approach: IRBL2}
\label{irbl2}

Importance Reweighting for Biquality Learning (IRBL) \cite{nodet2020importance} is a biquality learning algorithm designed to handle closed-set concept-drift. The algorithm is based on using two probabilistic classifiers: first, to estimate both concepts $\mathbb{P}_T(Y \mid X)$ and $\mathbb{P}_U(Y \mid X)$ and, second, using these classifiers' outputs to estimate the RND between both data distributions. In the particular case of label noise, especially instance dependent label noise, it has been shown to be the best approach experimentally on a wide variety of datasets.

We propose to adapt it to handle covariate shift by estimating the ratio $\frac{\mathbb{P}_T(X)}{\mathbb{P}_U(X)}$ by using a third probabilistic classifier based on Discriminative Learning \cite{bickel2007discriminative}. This algorithm works by defining a new supervised classification task by learning to predict if a sample is trusted or untrusted by only using its features. If there exists covariate shift between the datasets, the classifier should be able to discriminate between the two datasets.

Let's introduce $S$ as the new target: 
\begin{equation}
    s_i(x_i)=
\begin{cases}
  0, & \text{if}\ x_i \in D_U\\
  1, & \text{if}\ x_i \in D_T
\end{cases}
\label{s-equation}
\end{equation}

Estimating $\mathbb{P}(S \mid X)$ allows us to estimate $\frac{\mathbb{P}_T(X)}{\mathbb{P}_U(X)}$ directly without estimating both distributions:

\begin{equation}
\label{pdr-equation}
\begin{aligned}
    \frac{\mathbb{P}_T(X)}{\mathbb{P}_U(X)} = \frac{\mathbb{P}(X \mid S=1)}{\mathbb{P}(X \mid S=0)}
    &=\frac{\mathbb{P}(S=1 \mid X)\mathbb{P}(X)}{\mathbb{P}(S=1)} \times  \frac{\mathbb{P}(S=0)}{\mathbb{P}(S=0 \mid X)\mathbb{P}(X)} \\
    &=\frac{\mathbb{P}(S=1 \mid X)}{1 - \mathbb{P}(S=1 \mid X)} \times \frac{1 - \mathbb{P}(S=1)}{\mathbb{P}(S=1)}
\end{aligned}
\end{equation}

Combining Equation \ref{irbl-equation} and \ref{pdr-equation} :

\begin{equation}
    \label{irbl2-equation}
    \frac{\mathbb{P}_T(Y \mid X)\mathbb{P}_T(X)}{\mathbb{P}_U(Y \mid X)\mathbb{P}_U(X)}
    =\frac{\mathbb{P}_T(Y \mid X)}{\mathbb{P}_U(Y \mid X)}\times\frac{\mathbb{P}(S=1 \mid X)}{1 - \mathbb{P}(S=1 \mid X)}\times\frac{1 - \mathbb{P}(S=1)}{\mathbb{P}(S=1)}
\end{equation}

We propose to estimate Equation \ref{irbl2-equation} by learning probabilistic classifiers $f \in \mathcal{F}$ to estimate each of its terms. A probabilistic classifier $f_T$ is learned on $D_T$ to estimate $\mathbb{P}_T(Y \mid X)$, $f_U$ is learned on $D_U$ to estimate $\mathbb{P}_U(Y \mid X)$, and $f_S$ is learned on $\{(x,s(x)) \mid \forall x \in D_T \cup D_U\}$ to estimate $\mathbb{P}_U(S \mid X)$, leading to the following Algorithm \ref{irbl2-algo}.

\SetAlgoVlined% Similar to 'vlined' in the package load option
\RestyleAlgo{ruled}

\begin{algorithm}[!ht]
\small
\DontPrintSemicolon
\Indp
  \KwInput{Trusted Dataset $D_T$, Untrusted Dataset $D_U$, Probabilistic Classifier Family $\mathcal{F}$}
  Learn $f_U \in \mathcal{F}$ on $D_U$\;
  Learn $f_T \in \mathcal{F}$ on $D_T$\;
  Learn $f_S \in \mathcal{F}$ on $\{(x,s(x)) \mid \forall x \in D_T \cup D_U\}$\;
  \For{$(x_i,y_i) \in D_U$}
    { 
    	$\hat{\beta}(x_i,y_i) =  \frac{f_T(x_i)_{y_i}}{f_U(x_i)_{y_i}}\frac{f_S(x_i)_1}{\vert D_T \vert}\frac{\vert D_U \vert}{f_S(x_i)_0}$
    }
  \For{$(x_i,y_i) \in D_T$}  
    { 
    	$\hat{\beta}(x_i,y_i) = 1$
    }
  Learn $f \in \mathcal{F}$ on $D_T \cup D_U$ with weights $\hat{\beta}$ \;
  \KwOutput{$f$}
\caption{Importance Reweighting for Biquality Learning 2 (IRBL2)}
\label{irbl2-algo}
\end{algorithm}

\section[Second proposed approach: K-PDR]{Second proposed approach: $K$-PDR}
\label{kpdr}

$K$-DensityRatio ($K$-DR) \cite{fang2020rethinking} is an alternative approach to design a biquality learning algorithm able to handle distribution shift. The focus is made on the covariate shift between the two datasets. It handles the covariate shift in a class conditional fashion to deal with distribution shifts by using covariate shift correction once per class.

From Equation \ref{kdr-equation}, $K$-DR evaluates the ratio $\frac{\mathbb{P}_T(X \mid Y)}{\mathbb{P}_U(X \mid Y)}$ with density ratio estimation algorithms. To do so, it first samples data from the $X \mid Y$ distribution by selecting only samples from a given class $k \in [\![1,K]\!]$ in both datasets $D_T$ and $D_U$. Then, it uses density ratio estimation algorithms $e \in \mathcal{E}$ on these sub-datasets to estimate $\frac{\mathbb{P}_T(X \mid Y=k)}{\mathbb{P}_U(X \mid Y=k)}$ independently $k$ times. The class priors $\mathbb{P}_T(Y)$ and $\mathbb{P}_U(Y)$ are estimated empirically from both training sets. See Algorithm \ref{kdr-algorithm}.

\begin{algorithm}[!ht]
\small
\DontPrintSemicolon
\Indp
    \KwInput{Trusted Dataset $D_T$, Untrusted Dataset $D_U$, Density Ratio Estimator Family $\mathcal{E}$, Probabilistic Classifier Familiy $\mathcal{F}$}
    \For{$k \in [\![1,K]\!]$}
    {
        Let $D^k_T = \{\forall (x,y) \in D_T  \mid  y=k\}$\;
        Let $D^k_U = \{\forall (x,y) \in D_U  \mid  y=k\}$\;
        Learn $e^k \in \mathcal{E}$ on $D^k_T$ and $D^k_U$\;
    }
    \For{$(x_i,y_i) \in D_U$}
    {
        $\hat{\beta}(x_i,y_i) = e^{y_i}(x_i) \frac{ \vert D^{y_i}_T \vert }{ \vert D_T \vert } \frac{ \vert D_U \vert }{ \vert D^{y_i}_U \vert }$
    }
    \For{$(x_i,y_i) \in D_T$}  
    { 
    	$\hat{\beta}(x_i,y_i) = 1$
    }
    Learn $f \in \mathcal{F}$ on $D_T \cup D_U$ with weights $\hat{\beta}$\;
    \KwOutput{$f$}
\caption{$K$-DensityRatio ($K$-DR)}
\label{kdr-algorithm}
\end{algorithm}

In \cite{fang2020rethinking} Kernel Mean Matching (KMM) \cite{huang2007correcting,gretton2009covariate} has been used as the Density Ratio algorithm $e$ to handle covariate shift. Empirically, KMM is an algorithm that matches with quadratic programming \cite{wright1999continuous} the mean of both datasets in a feature space induced by a kernel $k$ on the domain $\mathcal{X}\times\mathcal{X}$:

\begin{equation}
\label{kmm-equation}
\begin{aligned}
\min_{\beta_i} \quad & \left\| \frac{1}{ \vert D_U \vert }\sum_{i=0}^{ \vert D_U \vert }\beta_i\Phi(x_i) - \frac{1}{ \vert D_T \vert }\sum_{i=0}^{ \vert D_T \vert }\Phi(x_i) \right\| _\mathcal{H}\\
\textrm{s.t.} \quad & 0\leq \beta_i \leq B \\ 
& \left\vert \frac{1}{ \vert D_U \vert }\sum_{i=0}^{ \vert D_U \vert }\beta_i -1 \right\vert  < \epsilon
\end{aligned}
\end{equation}
where $\Phi:\mathcal{X}\to\mathcal{H}$ denotes the canonical feature map, $\mathcal{H}$ is the reproducing kernel Hilbert space induced by the kernel $k$, $\ \mid \cdot\ \mid _\mathcal{H}$ is the norm on $\mathcal{H}$ and $B$ and $\epsilon$ are regularization and normalization constraints.

As such, KMM is a parametric algorithm based on kernels. We propose to use instead a probabilistic classifier to handle covariate shift, in the same fashion as in Equations \ref{s-equation} and \ref{pdr-equation} to make a non-parametric version of $K$-DR as shown in Equation \ref{kpdr-equation}.

\begin{equation}
\label{kpdr-equation}
    \begin{aligned}
    \frac{\mathbb{P}_T(X \mid Y)\mathbb{P}_T(Y)}{\mathbb{P}_U(X \mid Y)\mathbb{P}_U(Y)}
    &= \frac{\mathbb{P}(X \mid Y,S=1)}{\mathbb{P}(X \mid Y,S=0)}\times \frac{\mathbb{P}(Y \mid S=1)}{\mathbb{P}(Y \mid S=0)}\\
    &= \frac{\mathbb{P}(S=1 \mid X,Y)\mathbb{P}(X,Y)}{\mathbb{P}(Y \mid S=1)\mathbb{P}(S=1)}\times\frac{\mathbb{P}(Y \mid S=0)\mathbb{P}(S=0)}{\mathbb{P}(S=0 \mid X,Y)\mathbb{P}(X,Y)}\\
    &\times\frac{\mathbb{P}(Y \mid S=1)}{\mathbb{P}(Y \mid S=0)}\\
    &= \frac{\mathbb{P}(S=1 \mid X,Y)}{1 - \mathbb{P}(S=1 \mid X,Y)}\times\frac{1 -\mathbb{P}(S=1)}{\mathbb{P}(S=1)}
    \end{aligned}
    \tag{8.b}
\end{equation}

The main advantage of the non-parametric approach is that it does not require assumptions about the data distribution, which may not be satisfied in many real-world datasets and could lead to poor performances. Moreover, the scalability of $K$-PDR is better than the scalability of $K$-KMM both in space and time complexity. $K$-PDR has $K$ times the same complexity as the complexity of learning the chosen probabilistic classifier, which is $\mathcal{O}(K\times\vert \mathcal{X} \vert \times \vert D_u^k \vert \log ( \vert D_u^k \vert)) $ for Decisions Trees \cite{cormen2022introduction}. Meanwhile, $K$-KMM memory complexity is $\mathcal{O}(\vert D_u^k \vert^2 + \vert D_u^k \vert \times \vert D_t^k \vert)$ to sequentially build matrices necessary for the quadratic program, and a worst-case time complexity of $\mathcal{O}(K\times\vert D_u^k \vert^3)$ to solve the quadratic program \cite{ye1989extension}. Finally, the proposed approach is even more flexible than the previous one, as any family of machine learning classifiers could be used instead of kernels. This leads to Algorithm \ref{kpdr-algorithm}.

\begin{algorithm}[h]
\small
\DontPrintSemicolon
\Indp
    \KwInput{Trusted Dataset $D_T$, Untrusted Dataset $D_U$, Probabilistic Classifier Familiy $\mathcal{F}$}
    \For{$k \in [\![1,K]\!]$}
    {
        Let $D^k_T = \{\forall (x,y) \in D_T  \mid  y=k\}$\;
        Let $D^k_U = \{\forall (x,y) \in D_U  \mid  y=k\}$\;
        Learn $f^k_S \in \mathcal{F}$ on $\{(x,s(x)) \mid  \forall x \in D^k_T \cup D^k_U\}$\;
    }
    \For{$(x_i,y_i) \in D_U$}
    {
        $\hat{\beta}(x_i,y_i) = \frac{f_S^{y_i}(x_i)_1}{f_S^{y_i}(x_i)_0}\frac{ \vert D_U \vert }{ \vert D_T \vert }$
    }
    \For{$(x_i,y_i) \in D_T$}  
    { 
    	$\hat{\beta}(x_i,y_i) = 1$
    }
  Learn $f \in \mathcal{F}$ on $D_T \cup D_U$ with weights $\hat{\beta}$ \;
    \KwOutput{$f$}
\caption{$K$-Probabilistic Density Ratio ($K$-PDR)}
\label{kpdr-algorithm}
\end{algorithm}

\section{Experiments}
\label{experiments}

Benchmarking biquality learning algorithms means evaluating their efficiency and resilience on both dataset shifts and weaknesses of supervision in a joint manner. Introducing these corruptions synthetically in usual public multi-class classification datasets allows a fine-grained and controlled evaluation of these algorithms. 

From Equation \ref{distribution-shift-equation}, introducing distribution shift can be done in four ways: by introducing covariate shift, concept drift, class-conditional shift, or prior shift. Especially modifying both concept drift and covariate shift or class-conditional shift and prior shift at the same time leads to particularly complex distribution shifts. Table \ref{shifts-table} sums up the hierarchy of distribution shift sources.

\begin{table}[!h]
\centering
\caption{Hierarchy of Distribution Shift sources}
\label{shifts-table}
\small	
\begin{tabular}{|c|c|c|c|}
\hline
\multicolumn{4}{|c|}{Distribution Shift}                                                                  \\
\multicolumn{4}{|c|}{$\mathbb{P}(X,Y)$}                                                                  \\ \hline
\multicolumn{2}{|c|}{\multirow{2}{*}{$\mathbb{P}(Y \mid X)\mathbb{P}(X)$}} & \multicolumn{2}{c|}{\multirow{2}{*}{$\mathbb{P}(X \mid Y)\mathbb{P}(Y)$}}    \\ 
\multicolumn{2}{|c|}{} & \multicolumn{2}{c|}{}\\ \hline
\multicolumn{1}{|c|}{Concept Drift} & \multicolumn{1}{c|}{Covariate Shift} & \multicolumn{1}{c|}{Class-Conditional Shift} & \multicolumn{1}{c|}{Prior Shift} \\
\multicolumn{1}{|c|}{$\mathbb{P}(Y \mid X)$} & \multicolumn{1}{c|}{$\mathbb{P}(X)$} & \multicolumn{1}{c|}{$\mathbb{P}(X \mid Y)$} & \multicolumn{1}{c|}{$\mathbb{P}(Y)$} \\ \hline
\end{tabular}
\end{table}

We chose two methods to synthetically generate distribution shifts, one in each sub-tree of distribution shift sources: concept drift and class-conditional shift. We propose two novel ways to generate such shifts in real-world datasets that are detailed in the following Subsection \ref{concept-drift} and Subsection \ref{class-conditional-shift}.

\subsection{Concept Drift}
\label{concept-drift}

Concept drift corresponds to changes in the decision boundary $\mathbb{P}(Y\mid X)$ of a classifier in some parts of the feature space $\mathcal{X}$. 

To synthetically generate concept drift in real-world datasets, we propose to model the feature space with a Decision Tree classifier \cite{breiman1984classification} learned on the original dataset $D$, such that each leaf of the decision tree will correspond to a patch of the dataset grid. By restricting the minimum samples per leaf in the decision tree, here set as 10\% of the dataset per class, we will be able to control the finesse of the grid. 

Then in each leaf of the decision tree, we are going to change the class distribution, essentially by generating noisy labels $\tilde{Y}$ from clean labels $Y$ thanks to a transition matrix $\mathbf{T}$, such that $\forall (i,j) \in [\![1,K]\!]^2, \mathbf{T}_{i,j} = \mathbb{P}(\tilde{Y}=j\mid Y=i)$. We chose to restrict the transition matrix to a permutation matrix $\mathbf{P}$, meaning that each clean class will be associated in a bijective way with a noisy class, different from the clean class. These permutation matrices $\mathbf{P}$ will be generated randomly once per dataset. For example, for three-class classification, one of these permutation matrices could be the following:
\begin{equation*}
    \mathbf{P} = \begin{pmatrix}
0 & 1 & 0\\
0 & 0 & 1\\
1 & 0 & 0
\end{pmatrix}
\end{equation*}

To decide how much concept drift will occur, we rank the leaves in the decision tree by their purity, starting from the purest leaves, and chose enough leaves such that $r\%$ of the untrusted dataset fall in these leaves. These samples will be given a noisy label, and the rest of the samples will be left untouched. 

The benefits of this methodology are twofold:
\begin{itemize}
    \item Having all samples assigned the same noisy label given their class in a whole subspace of the feature space will avoid the class overlap usually created with uniform (completely at random) label noise.
    \item Choosing to add label noise in the purest leaves of the decision tree first creates new patterns in otherwise completely clean and easy parts of the dataset.
\end{itemize}

\subsection{Class-Conditional Shift}
\label{class-conditional-shift}

Class-conditional shift corresponds to changes in the feature distribution $\mathbb{P}(X)$, with different changes per class $\mathbb{P}(X\mid Y)$.

To synthetically generate class-conditional shift in a real-world dataset, we propose to split the original dataset $D$ into sub-datasets $D^k$, such that $\forall k \in [\![1,K]\!], D^k=\{(x,y)\in D\mid y=k\}$, where $D^k$ corresponds to samples from the same class $k$ out of all the classes $K$.

For each sub-dataset, we train a K-Means \cite{lloyd1982least} clustering to learn a division of the feature space $\mathcal{X}$ given the class $k$. The number of clusters is chosen by cross-validation to maximize the average silhouette score \cite{ROUSSEEUW198753}.

Then we sub-sample some of these clusters to modify the class-conditional distribution $\mathbb{P}(X \mid Y=k)$. We sort the clusters by their size and separate them in two groups of equal size. The group formed by the smallest clusters will be subsampled according to a subsampling ratio $\rho$. The group formed by the biggest clusters will be left untouched. For example, if the number of clusters for a class is 4 and $\rho=10$, the two smallest clusters will be subsampled by a ratio of $10$, the two biggests will not be subsampled.

A higher $\rho$ will lead to more data being missing in the untrusted dataset, especially samples in sparse regions of the feature space, represented by small clusters.

The final size of the untrusted dataset relative to the non-sub-sampled untrusted dataset is reported for each scenario in Table \ref{table-conversion-subsampling} in the Appendices.

\subsection{Illustration on a Toy Dataset} 

We propose to first see the experimental protocol on a toy dataset with two features to visualize better how the corruption mechanisms will act. We chose the two moons dataset, composed of two croissant-shaped classes that are almost linearly separable.

\begin{figure}[!h]
\centering
\includegraphics[width=0.35\linewidth]{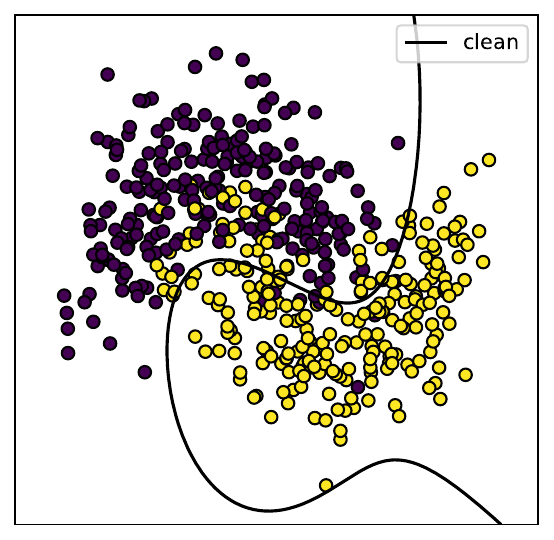}
\caption{Two moons dataset with the decision boundary of a Support Vector Machine classifier using a polynomial kernel.}
\label{two-moons}
\end{figure}

Figure \ref{two-moons} shows that a Support Vector classifier \cite{boser1992training} with a Polynomial kernel of degree 3 can adequately separate the two classes.

Now we can apply our synthetic corruptions to this dataset and see the effect on the decision boundary of the Support Vector classifier.

\begin{figure}[!h]
\centering
\begin{tabular}{cc}
    \begin{subfigure}[b]{0.35\linewidth}
         \includegraphics[width=\textwidth]{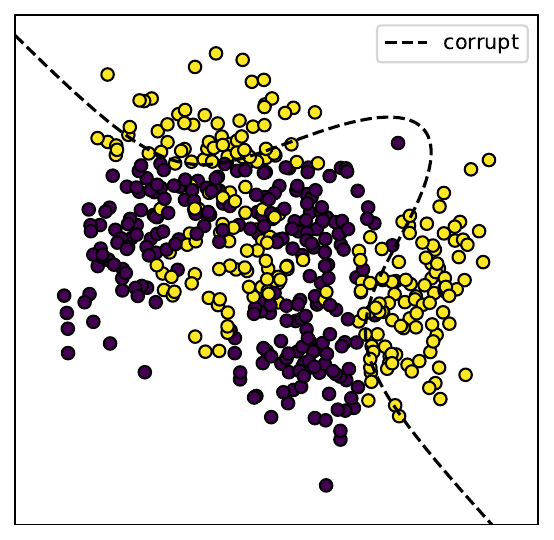}
        \caption{Concept drift}
    \end{subfigure}
 &
    \begin{subfigure}[b]{0.35\linewidth}
         \includegraphics[width=\textwidth]{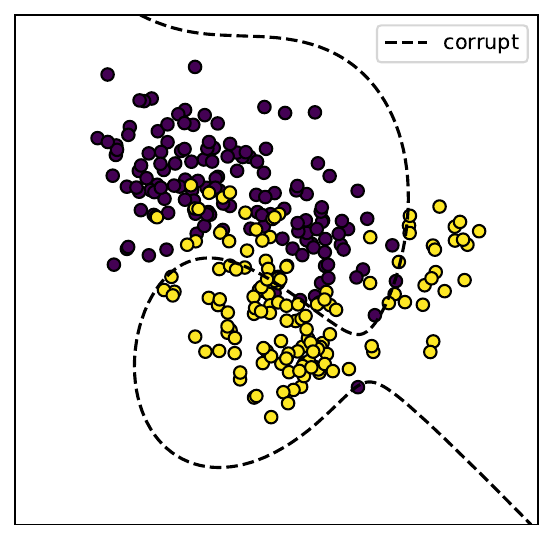}
        \caption{Class-conditional shift}
    \end{subfigure}
\\
\end{tabular}
\caption{Corrupted two moons dataset with the decision boundary of a Support Vector Machine classifier using a polynomial kernel.}
\label{two-moons-corrupted}
\end{figure} 

The proposed synthetic corruptions affect the learned decision boundary of the Support Vector classifier but each in their own way. The concept drift dramatically changes the decision boundary, and the classifier cannot recognize the two original croissants illustrated in Figure \ref{two-moons-corrupted}. Instead, it emphasizes the new class patterns created with the noisy patches. The class-conditional shift, however, does not inherently change the shape of the decision boundary. We can still see that the classifier somewhat separates the two croissants. However, the margin is way tighter on the yellow croissant. This classifier would fail to classify some data points that would be considered out-of-distribution for the corrupted dataset but completely normal in the original dataset.

Now that we verified that the synthetic corruptions produce the expected effects on classifiers, we can look at how the previous method of the state-of-the-art can correct these corrupted datasets, especially IRBL and PDR that are expected to work only on concept drift or covariate shift, and our two proposed algorithm, IRBL2, and $K$-PDR, that should work on distribution shifts. We propose to keep 5\% of the origin two moons dataset as trusted data that will be left untouched and corrupt the rest of the dataset with both corruptions.

\begin{figure}[!h]
\centering
\begin{tabular}{ccc}
    \begin{subfigure}[b]{0.30\linewidth}
         \includegraphics[width=\textwidth]{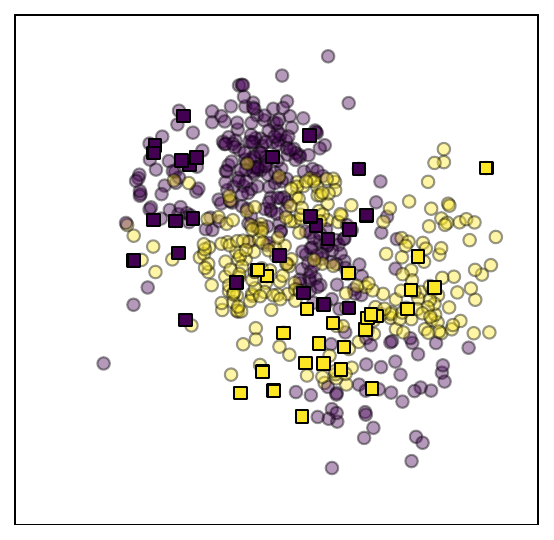}
        \caption{No Correction}
    \end{subfigure}
 &
    \begin{subfigure}[b]{0.30\linewidth}
         \includegraphics[width=\textwidth]{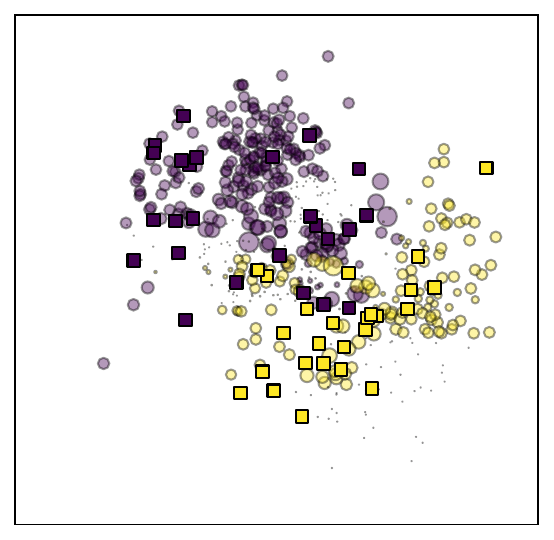}
        \caption{IRBL}
    \end{subfigure}
    &
    \begin{subfigure}[b]{0.30\linewidth}
         \includegraphics[width=\textwidth]{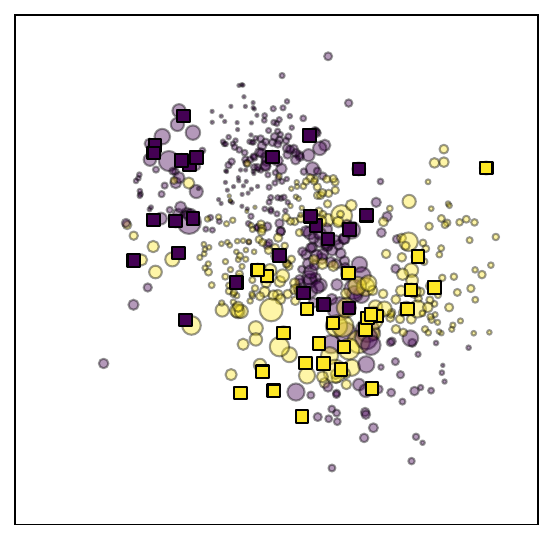}
        \caption{PDR}
    \end{subfigure}
    \\
    &
    \begin{subfigure}[b]{0.30\linewidth}
         \includegraphics[width=\textwidth]{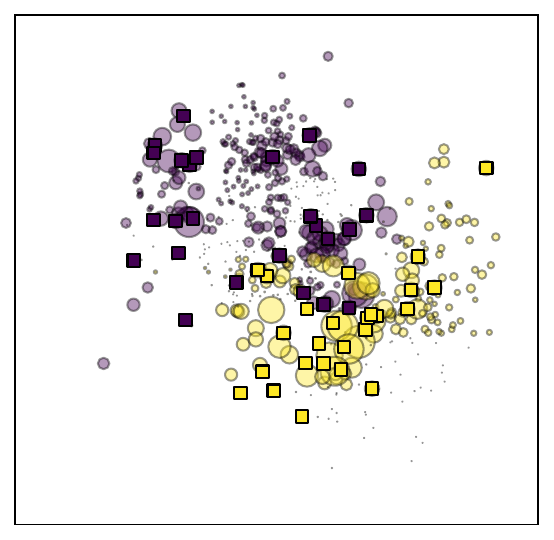}
        \caption{IRBL2}
    \end{subfigure}
    &
    \begin{subfigure}[b]{0.30\linewidth}
         \includegraphics[width=\textwidth]{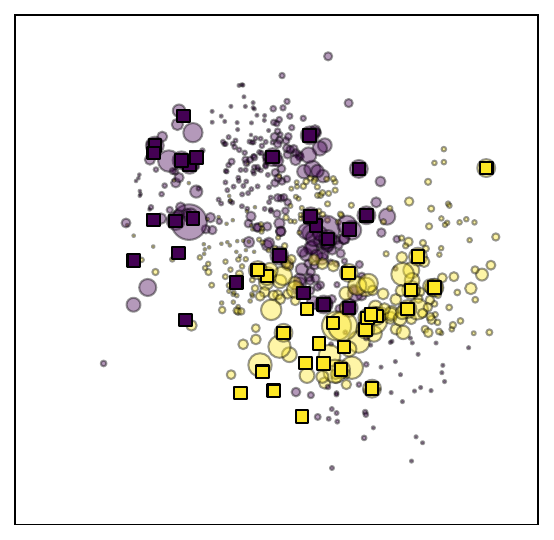}
        \caption{$K$-PDR}
    \end{subfigure}
\\
\end{tabular}
\caption{Trusted and untrusted two moons dataset. Trusted data points are represented with a $\square$ marker and untrusted data points are represented with a $\bigcirc$ marker and with a more transparent color. The markers' sizes are proportional to the weights given by the corresponding algorithm.}
\label{two-moons-corrected}
\end{figure}

In Figure \ref{two-moons-corrected}, we can see the limitations of IRBL and PDR. IRBL can detect samples belonging to the wrong part of the feature space given their label but cannot take into account the uncertainty of some patches of the feature space that are not present in the trusted dataset. For example, in the top part of the dataset, there is a zone where no trusted points exist for the purple moon. IRBL do not takes that into account and re-weights equally all these, as long as they are from the correct color, purple. Meanwhile, PDR does re-weight the samples differently depending if they belong to the trusted distribution, but indifferently depending on the sample color. 

In the second part of Figure \ref{two-moons-corrected}, we can observe that IRBL2 and $K$-PDR can handle both cases and, thus, distribution shift. Their re-weighting schemes are reasonably similar, and to assess which one is better, we need to benchmark them on a broader range of real-world datasets.

\subsection{Datasets}

We randomly picked supervised classification datasets, see Table \ref{datasets}, from different sources: UCI \cite{Dua2019}, libsvm \cite{chang2011libsvm}, active learning challenge \cite{guyon2010dataset} and openML \cite{10.1145/2641190.2641198}. A part of these datasets comes from past challenges in active learning, where high performances with a low number of labeled samples have proved challenging to obtain, which makes leveraging the untrusted dataset necessary. With this choice of datasets, an extensive range of class ratios, number of classes, number of features, and dataset sizes are covered.

\begin{table}[!h]
\caption{Multi-class classification datasets used for the evaluation. Columns: number of samples ($ \vert D \vert $), number of features ($ \vert \mathcal{X} \vert $), number of classes ($ \vert \mathcal{Y} \vert $), and ratio of the number of samples from the minority class over the number of samples from the majority class (min).}
\centering
\footnotesize
\setlength{\tabcolsep}{2pt}
\begin{tabular}{llrrr}
\toprule
Datasets & $ \vert D \vert $ & $ \vert \mathcal{X} \vert $ & $ \vert \mathcal{Y} \vert $ & min \\
\midrule
eeg & 15K & 14 & 2 & 0.81 \\
ibn-sina & 20.7K & 92 & 2 & 0.61 \\
zebra & 61.5K & 154 & 2 & 0.05 \\
musk & 6.6K & 167 & 2 & 0.18 \\
phishing & 11.1K & 30 & 2 & 0.80 \\
spam & 4.6K & 57 & 2 & 0.65 \\
ijcnn1 & 192K & 22 & 2 & 0.11 \\
hiva & 42.7K & 1617 & 2 & 0.04 \\
svmguide3 & 1.24K & 22 & 2 & 0.31 \\
web & 37K & 123 & 2 & 0.32 \\
mushroom & 8.12K & 22 & 2 & 0.93 \\
skin-seg & 245K & 3 & 2 & 0.26 \\
mozilla4 & 15.5K & 5 & 2 & 0.49 \\
electricity & 45.3K & 8 & 2 & 0.74 \\
bank-marketing & 45.2K & 16 & 2 & 0.13 \\
magic-telescope & 19K & 10 & 2 & 0.54 \\
phoeneme & 5.4K & 5 & 2 & 0.42 \\
poker & 1.03M & 10 & 2 & 1.00 \\
\bottomrule
\end{tabular}
\setlength{\tabcolsep}{2pt}
\begin{tabular}{lrrrr}
\toprule
Datasets & $ \vert D \vert $ & $ \vert \mathcal{X} \vert $  & $ \vert \mathcal{Y} \vert $ & min \\
\midrule
ad & 3.28K & 1558 & 2 & 0.16 \\
nomao & 34.5K & 118 & 2 & 0.40 \\
click & 39.9K & 9 & 2 & 0.20 \\
jm1 & 10.9K & 21 & 2 & 0.24 \\
pendigits & 11K & 16 & 10 & 0.92 \\
japanese-vowels & 9.96K & 14 & 9 & 0.48 \\
gas-drift & 13.9K & 128 & 6 & 0.55 \\
connect-4 & 67.6K & 42 & 3 & 0.15 \\
satimage & 6.43K & 36 & 6 & 0.41 \\
dna & 3.19K & 180 & 3 & 0.46 \\
first-ord-theo-prov & 6.12K & 51 & 6 & 0.19 \\
artificial-chars & 10.2K & 7 & 10 & 0.42 \\
splice & 3.19K & 60 & 3 & 0.46 \\
har & 10.3K & 561 & 6 & 0.72 \\
usps & 9.3K & 256 & 10 & 0.46 \\
mnist & 70K & 784 & 10 & 0.80 \\
letter & 20K & 16 & 26 & 0.90 \\
isolet & 7.8K & 617 & 26 & 0.99 \\
\bottomrule
\end{tabular}
\label{datasets}
\end{table}

Each dataset is first jumbled, then split in a stratified fashion with 80\% of samples used for training set, and 20\% used for the test set. 

Each training set is again split in a stratified fashion into a trusted and untrusted set with a ratio of trusted $p$. To find a ratio of trusted data that is coherent between all datasets, we compute a learning curve on the training set and choose a ratio of trusted data such that the classifier's performance is equal to $p\%$ of the classifier learned on the entire training set. The actual ratio of trusted data for each dataset is provided in Table \ref{table-conversion-trusted} in the Appendices.

Finally, we synthetically corrupt the untrusted set with the methods described earlier in this Section.

\subsection{Competitors}

We compare IRBL2 and $K$-PDR against multiple state-of-the-art competitors and baselines :
\begin{itemize}
    \item $K$-KMM, the original version of $K$-DR using KMM as proposed in \cite{fang2020rethinking};
    \item IRBL \cite{nodet2020importance}, which is IRBL2 without covariate shift correction;
    \item PDR \cite{bickel2007discriminative} the covariate-shift only baseline;
    \item Trusted-Only baseline, when the model is learned using only the trusted dataset;
    \item No-Correction baseline, when the model is learned on both datasets without correction applied.
\end{itemize}

For every competitors, we use the same probabilistic classifier family, histogram-based gradient boosting (HGBT) trees \cite{NIPS2017_6449f44a} from Scikit-Learn \cite{scikit-learn} with their default hyperparameters, each time an algorithm required a base classifier. When it required a well-calibrated classifier, the HGBT trees where calibrated thanks to an Isotonic Regression \cite{zadrozny2001calibration} from Scikit-Learn using Zadrozny's heuristic \cite{zadrozny2002transforming} for multiclass calibration.

For KMM and $K$-KMM we use the Radial Basis Function (RBF) kernel \cite{vert2004primer} with the default $\gamma = \frac{1}{\vert \mathcal{X} \vert}$ value from Scikit-Learn. In order to scale KMM to the bigger datasets, we used an ensembling version \cite{miao2015ensemble} of the original KMM algorithm with a batch size of 100.

\section{Results}
\label{results}

We conducted the previously defined experiments of the proposed datasets with the label noise strength $r$ varying from $0\%$ to $50\%$ of the dataset and the maximum cluster subsampling $\rho$ from $1$ to $100$. We also tested three different values for the ratio of trusted data $p$ equals to $25\%$, $50\%$, and $75\%$. The predictive performance we used in these experiments is Cohen's kappa coefficient $\kappa$ \cite{cohen1960coefficient}. It is a measure of agreement between assignments relative to the agreement between two random assignments and is fitted to evaluate classifier performance in unbalanced settings. Before looking at the whole grid of experiments over $r$ and $\rho$, we can look at the two axes independently to independently analyze the impact of concept drift and class-conditional shift.

The primary metric to analyze to assess the efficiency of biquality learning algorithms is the evolution of the predictive performance $\kappa$ given the corruption strength ($\rho$ or $r$). For example, having a constant predictive performance given the corruption strength means being robust to this corruption. 

We propose to illustrate this function by plotting the averaged predictive performance over all datasets for all tested corruption strength and ratio of trusted data.

\begin{figure}[!h]
\centering
\begin{tabular}{ccc}
    \begin{subfigure}[b]{0.30\linewidth}
         \includegraphics[width=\textwidth]{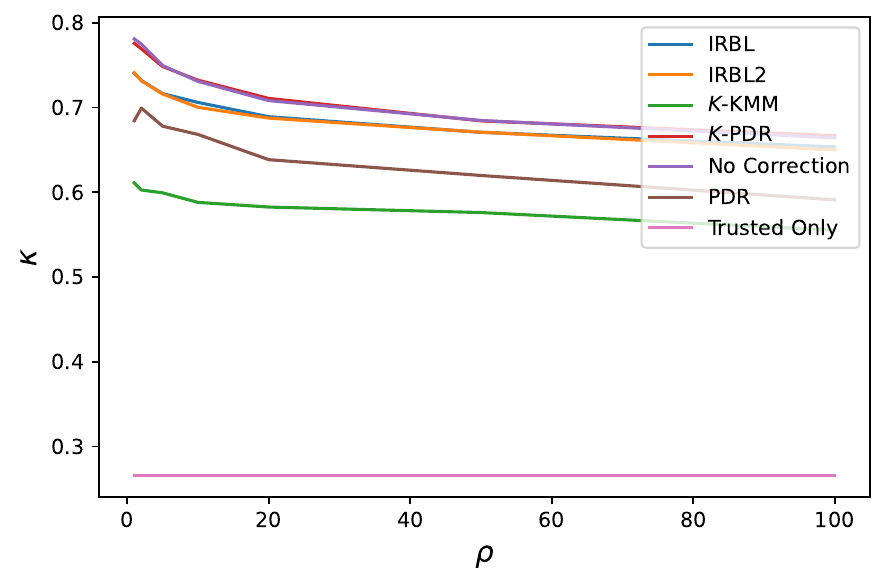}
        \caption{$r=0$, $p=0.25$}
    \end{subfigure}
 &
    \begin{subfigure}[b]{0.30\linewidth}
         \includegraphics[width=\textwidth]{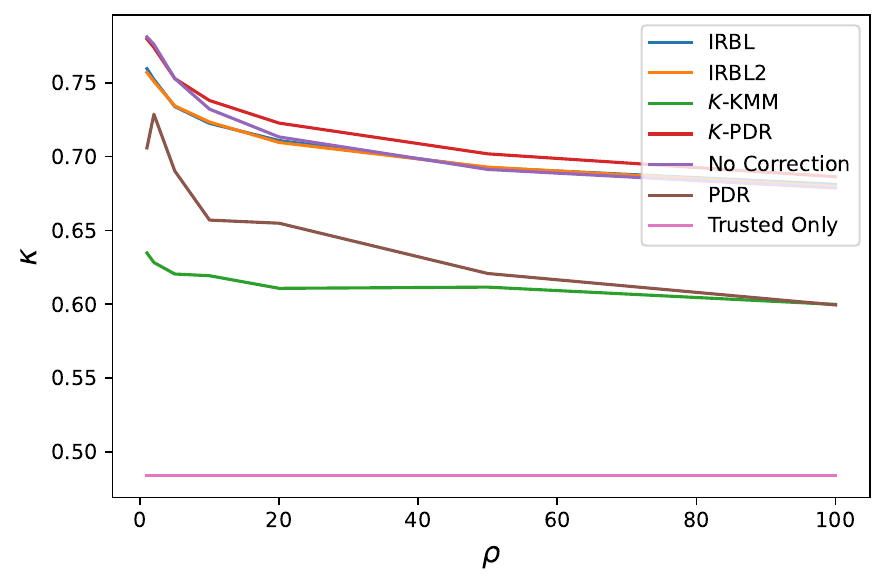}
        \caption{$r=0$, $p=0.5$}
    \end{subfigure}
    &
        \begin{subfigure}[b]{0.30\linewidth}
         \includegraphics[width=\textwidth]{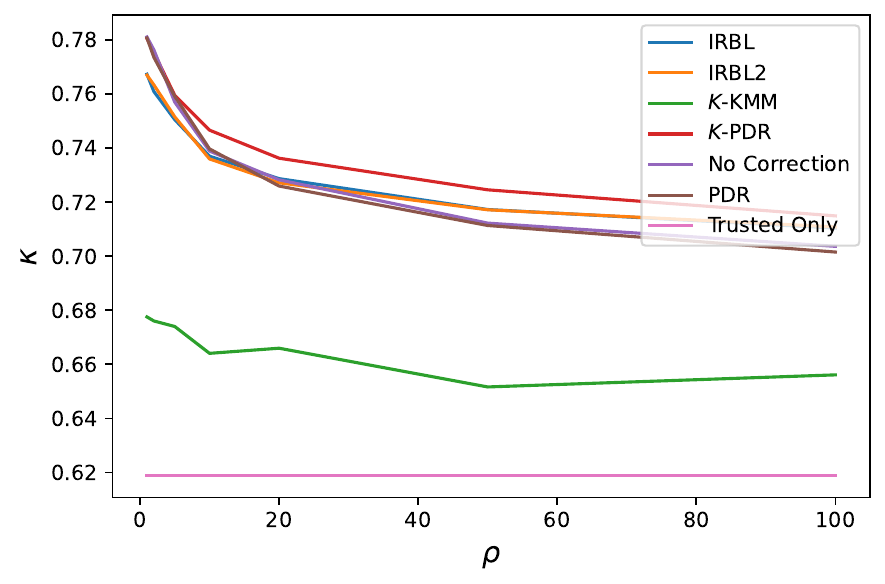}
        \caption{$r=0$, $p=0.75$}
    \end{subfigure}
    \\
    \begin{subfigure}[b]{0.30\linewidth}
         \includegraphics[width=\textwidth]{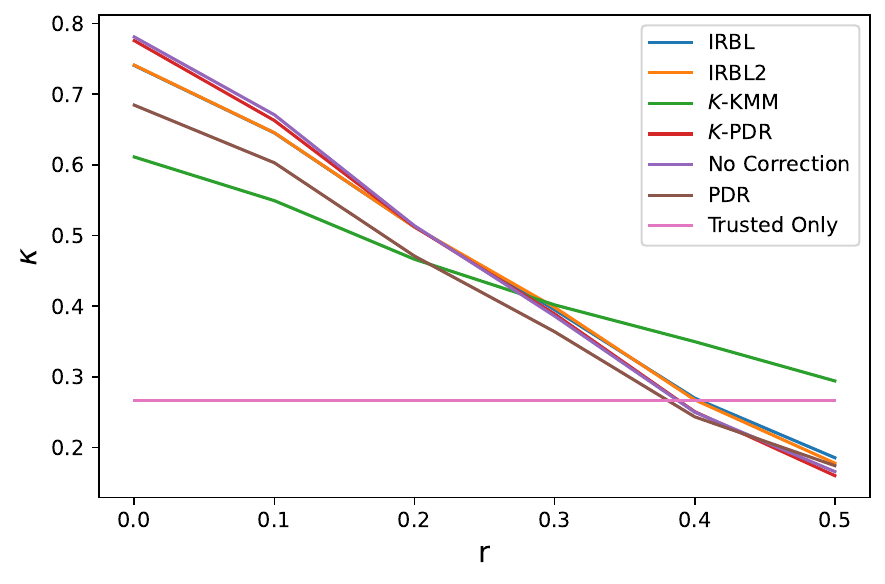}
        \caption{$\rho=0$, $p=0.25$}
    \end{subfigure}
 &
    \begin{subfigure}[b]{0.30\linewidth}
         \includegraphics[width=\textwidth]{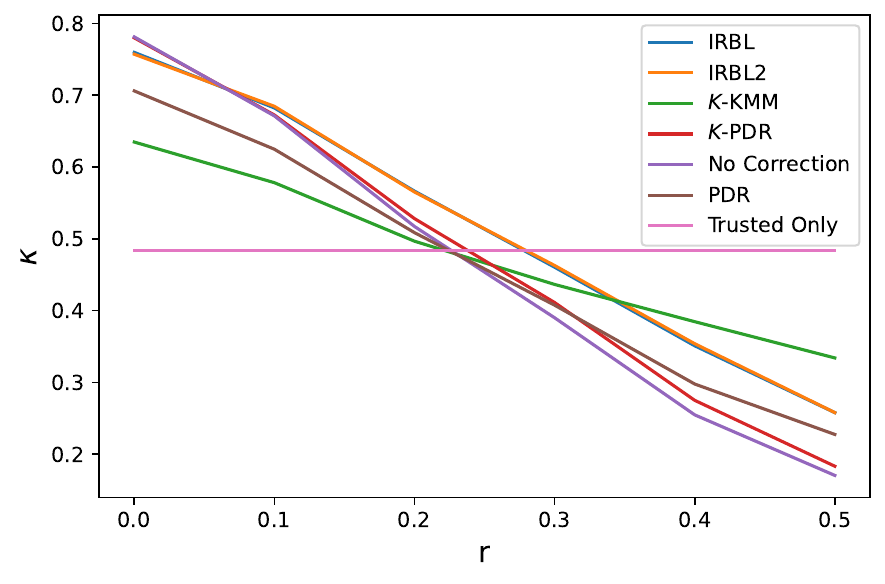}
        \caption{$\rho=0$, $p=0.5$}
    \end{subfigure}
    &
        \begin{subfigure}[b]{0.30\linewidth}
         \includegraphics[width=\textwidth]{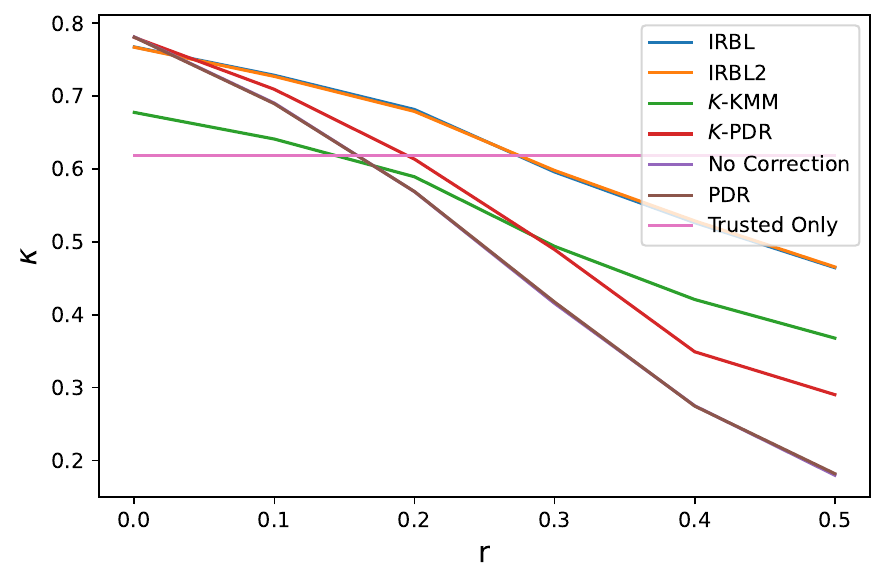}
        \caption{$\rho=0$, $p=0.75$}
    \end{subfigure}
    \\
\end{tabular}
\caption{Average Cohen's kappa $\kappa$ with different corruption strengths (noise ratio $r$, or cluster imbalance $\rho$) and ratio of trusted data $p$ over all datasets. The first row corresponds to experiments without noise ($r=0$), and the second row corresponds to experiments without cluster imbalance ($\rho=0$). Each column corresponds to a different case of ratio of trusted data ($p=0.25$, $p=0.5$, $p=0.75$).}
\label{avg-error-curves}
\end{figure}

Figure \ref{avg-error-curves} shows that with very few trusted data, $p=0.25$, biquality-learning algorithms have trouble differentiating themselves from one another and with baselines, showing their limitations on low trusted data regimes. However, $p=0.25$ leads to having around only thousandths of the data as trusted, which we consider to be a particularly pessimistic scenario in practice. With more trusted data, some competitors start to pull apart, especially $K$-PDR on class-conditional shift and IRBL and IRBL2 on concept drift.

In order to summarize these plots, we propose to compute the area under the curve (AUC) of the previous error curves, which is primarily a fancier average when the data points are not evenly spaced. This AUC is then normalized by the domain length we integrate the curve on.

\begin{table}
\caption{Averaged area under the Cohen's kappa $\kappa$ curve for all datasets.}
\centering
\begin{tabular}{rrlllllll}
\toprule
& p & IRBL & IRBL2 & $K$-KMM & $K$-PDR & No Corr. & PDR & Trusted \\
\midrule
\multirow{3}{*}{\rotatebox[origin=c]{90}{$r=0$}} & 0.25 & 0.676 & 0.675 & 0.575  & $\mathbf{0.694}$  & 0.693  & 0.624 & 0.266\\
& 0.50 & 0.699  & 0.698 & 0.610  & $\mathbf{0.709}$  & 0.700  & 0.630  & 0.484 \\
& 0.75 & 0.722  & 0.722  & 0.658 & $\mathbf{0.729}$  & 0.719  & 0.718 & 0.619 \\
\midrule
\multirow{3}{*}{\rotatebox[origin=c]{90}{$\rho=0$}}&  0.25 & 0.457 & 0.457 & 0.444 & 0.457 & $\mathbf{0.459}$ & 0.422 & 0.266 \\
& 0.50 & $\mathbf{0.515}$ & $\mathbf{0.515}$ & 0.476 & 0.474 & 0.462 & 0.461 & 0.484 \\
& 0.75 & $\mathbf{0.630}$ & $\mathbf{0.630}$ & 0.534 & 0.539 & 0.486 & 0.486 & 0.619\\
\bottomrule
\end{tabular}
\label{auc-table}
\end{table}

% \begin{table}
% \caption{Averaged area under the Cohen's kappa $\kappa$ curve for all datasets.}
% \centering
% \begin{tabular}{rrlllllll}
% \toprule
% & p & IRBL & IRBL2 & $K$-KMM & $K$-PDR & No Corr. & PDR & Trusted \\
% \midrule
% \multirow{3}{*}{\rotatebox[origin=c]{90}{$r=0$}} & 0.25 & 0.684 & 0.649 & 0.584 & 0.641 & $\mathbf{0.697}$ & 0.666 & 0.268 \\
% & 0.50 & 0.703 & 0.689 & 0.620 & 0.678 & $\mathbf{0.705}$ & 0.694 & 0.487 \\
% & 0.75 & $\mathbf{0.725}$ & 0.711 & 0.667 & 0.702 & 0.723 & 0.712 & 0.620 \\
% \midrule
% \multirow{3}{*}{\rotatebox[origin=c]{90}{$\rho=0$}}& 0.25 & $\mathbf{0.459}$ & 0.449 & 0.45 & 0.451 & 0.456 & 0.443 & 0.268 \\
% & 0.50 & 0.513 & 0.518 & 0.482 & $\mathbf{0.519}$ & 0.459 & 0.446 & 0.487 \\
% & 0.75 & 0.63 & $\mathbf{0.638}$ & 0.539 & 0.594 & 0.485 & 0.484 & 0.62 \\
% \bottomrule
% \end{tabular}
% \label{auc-table}
% \end{table}

Table \ref{auc-table} mostly confirmed what could be observed in Figure \ref{avg-error-curves}, but more particularly highlights the mere improvements brought by IRBL2 over IRBL with both corruptions.

\begin{figure}[!h]
\centering
\begin{tabular}{ccc}
    \begin{subfigure}[b]{0.28\linewidth}
         \includegraphics[width=\textwidth]{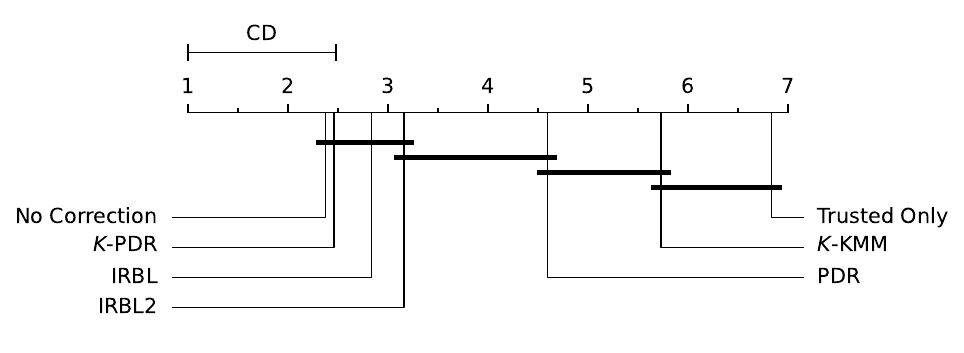}
        \caption{$r=0$, $p=0.25$}
    \end{subfigure}
 &
    \begin{subfigure}[b]{0.28\linewidth}
         \includegraphics[width=\textwidth]{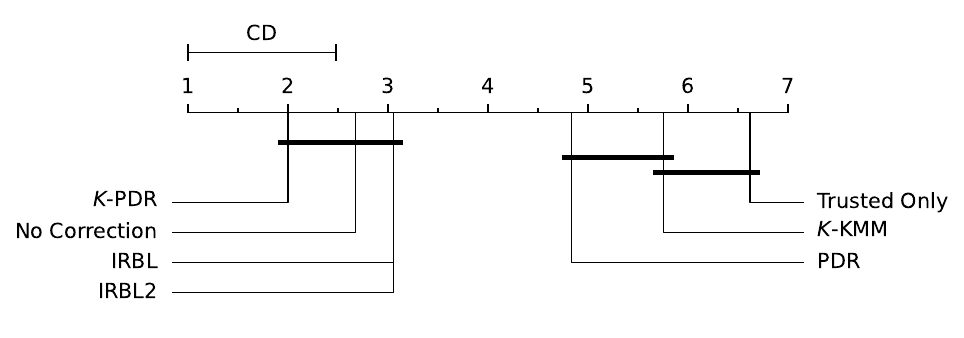}
        \caption{$r=0$, $p=0.5$}
    \end{subfigure}
    &
        \begin{subfigure}[b]{0.28\linewidth}
         \includegraphics[width=\textwidth]{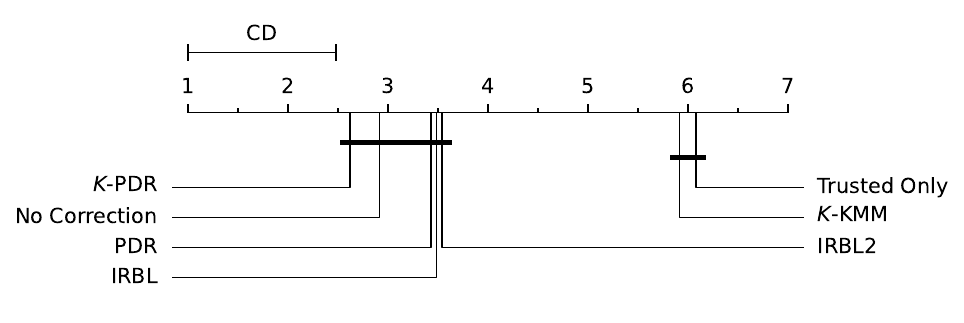}
        \caption{$r=0$, $p=0.75$}
    \end{subfigure}
    \\
    \begin{subfigure}[b]{0.28\linewidth}
         \includegraphics[width=\textwidth]{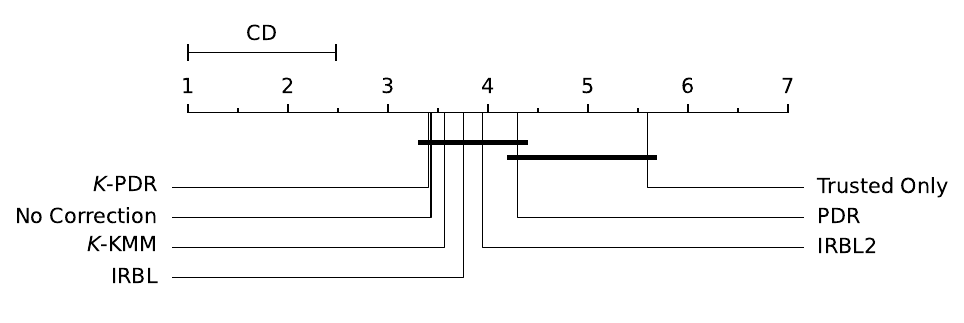}
        \caption{$\rho=0$, $p=0.25$}
    \end{subfigure}
 &
    \begin{subfigure}[b]{0.28\linewidth}
         \includegraphics[width=\textwidth]{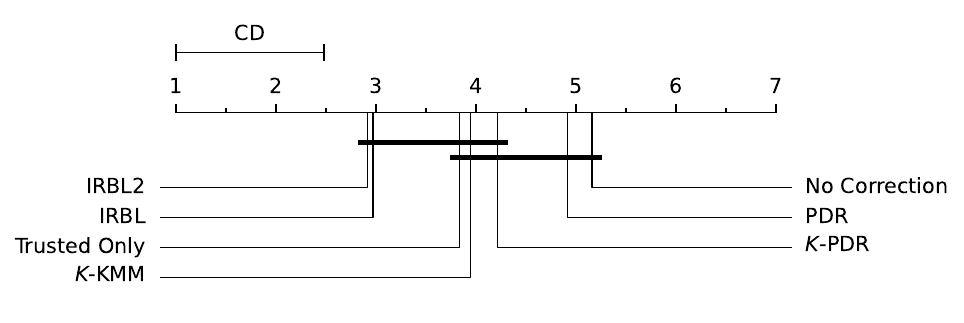}
        \caption{$\rho=0$, $p=0.5$}
    \end{subfigure}
    &
        \begin{subfigure}[b]{0.28\linewidth}
         \includegraphics[width=\textwidth]{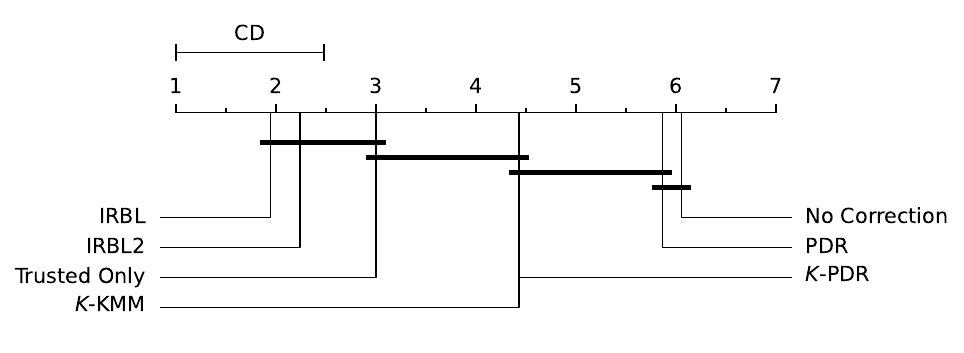}
        \caption{$\rho=0$, $p=0.75$}
    \end{subfigure}
    \\
\end{tabular}
\caption{Critical diagrams with different ratio of trusted data $p$ over all datasets. The first row corresponds to experiments without noise ($r=0$), and the second row corresponds to experiments without cluster imbalance ($\rho=0$). Each column corresponds to a different case of ratio of trusted data ($p=0.25$, $p=0.5$, $p=0.75$).}
\label{critical-diagrams}
\end{figure}

As the previous comparison are made using averaged predictive measures that might not be relatable from one dataset to the other, we propose in a second time to compute the average ranking instead thanks to critical diagrams presented in Figure \ref{critical-diagrams} as a more robust way to compare competitors.

The Nemenyi test \cite{Nemenyi62} is used to rank the approaches in terms of AUC of the predictive metric over the corruption strength. The Nemenyi test consists of two successive steps. First, the Friedman test is applied to the AUC of competing approaches to determine whether their overall performance is similar. Second, if not, the post-hoc test is applied to determine groups of approaches whose overall performance is significantly different from that of the other groups.

Figure \ref{critical-diagrams} confirms all previous observations. $K$-PDR seems to be the best approach on class-conditional noise but is still somewhat robust to label noise, especially at medium and high levels of trusted data ratio as its able to detach from PDR in terms of average rank in these situtations. IRBL and IRBL2 seem to be the best approaches for learning with label noise, even though they struggle with a low ratio of trusted data. However, IRBL2 struggles to improve IRBL performances on class-conditional shifts.

% Figure \ref{critical-diagrams} confirms all previous observations. $K$-PDR seems to be the best approach on class-conditional noise but is still somewhat robust to label noise, especially at medium and high levels of trusted data ratio. IRBL and IRBL2 seem to be the best approaches for learning with label noise, even though they struggle with a low ratio of trusted data. However, IRBL2 struggles to improve IRBL performances on class-conditional shifts. Additionally, $K$-KMM seems to be the most robust competitor for label noise at a very low ratio of trusted. Moreover, Figure \ref{critical-diagrams} shows that the breaking point where the predictive performance of the trusted only baseline wins over the predictive performance of IRBL and IRBL2 for $\rho=0$ seems to stabilize at $r=0.3$ for a growing ratio of trusted data.

Finally, to study the combined effect of both axes, we extend the experiments on a grid with varying strength of label noise $r$ and sub-sampling $\rho$.

\begin{figure}[h]
\centering
\captionsetup{justification=centering}
\begin{tabular}{ccc}
    \begin{subfigure}[b]{0.3\linewidth}
         \includegraphics[width=\textwidth]{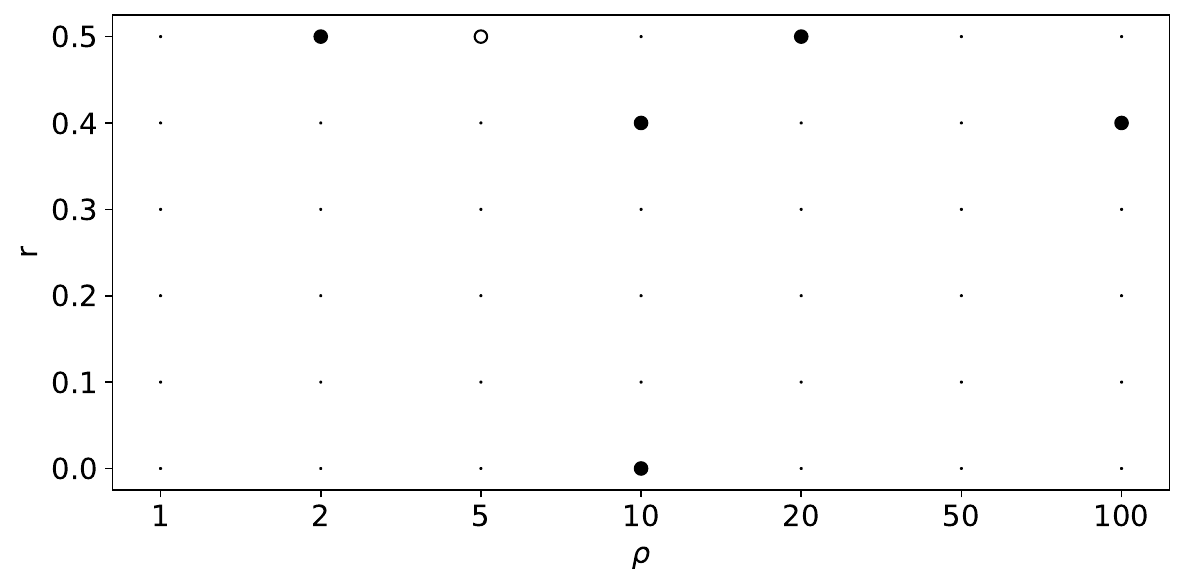}
        \caption{IRBL2 vs IRBL, $p=0.25$}
    \end{subfigure}
    &
    \begin{subfigure}[b]{0.3\linewidth}
         \includegraphics[width=\textwidth]{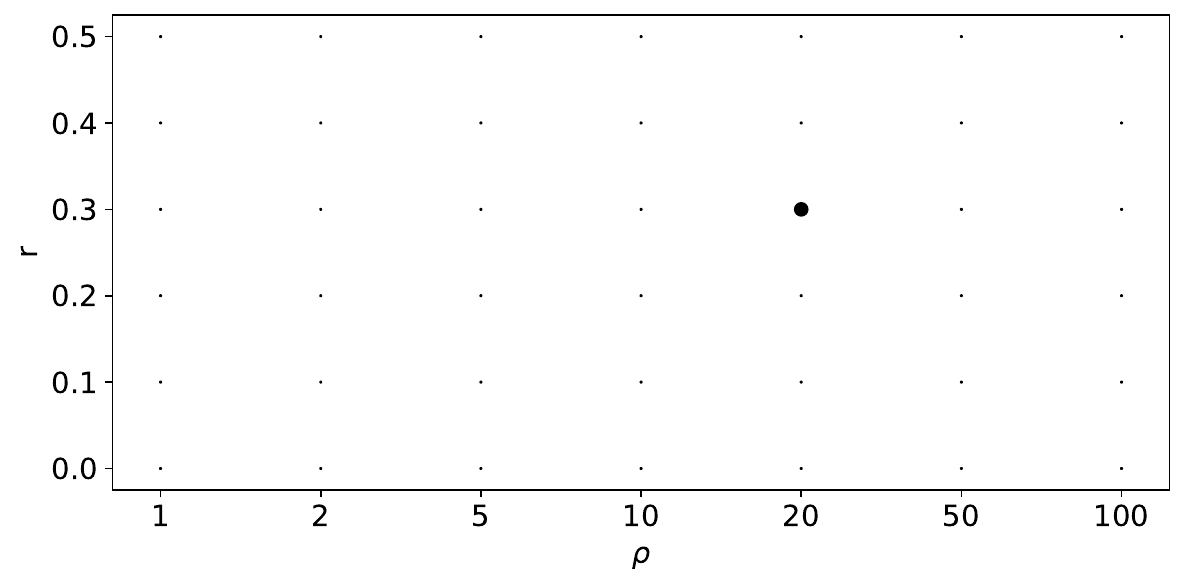}
        \caption{IRBL2 vs IRBL, $p=0.5$}
    \end{subfigure}
    &
    \begin{subfigure}[b]{0.3\linewidth}
         \includegraphics[width=\textwidth]{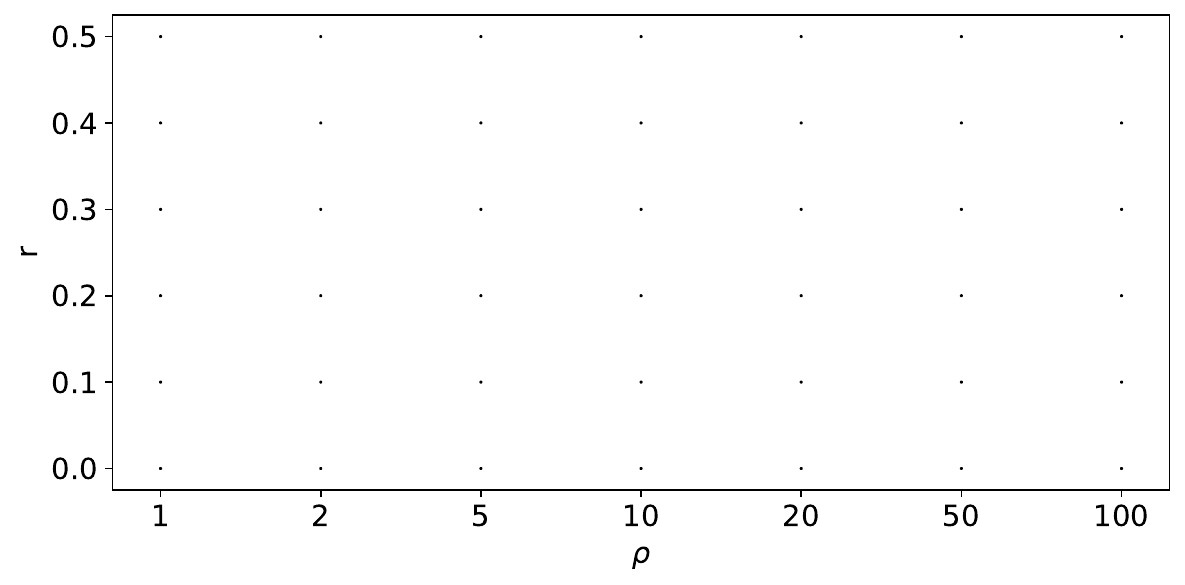}
        \caption{IRBL2 vs IRBL, $p=0.75$}
    \end{subfigure}
    \\
    \begin{subfigure}[b]{0.3\linewidth}
         \includegraphics[width=\textwidth]{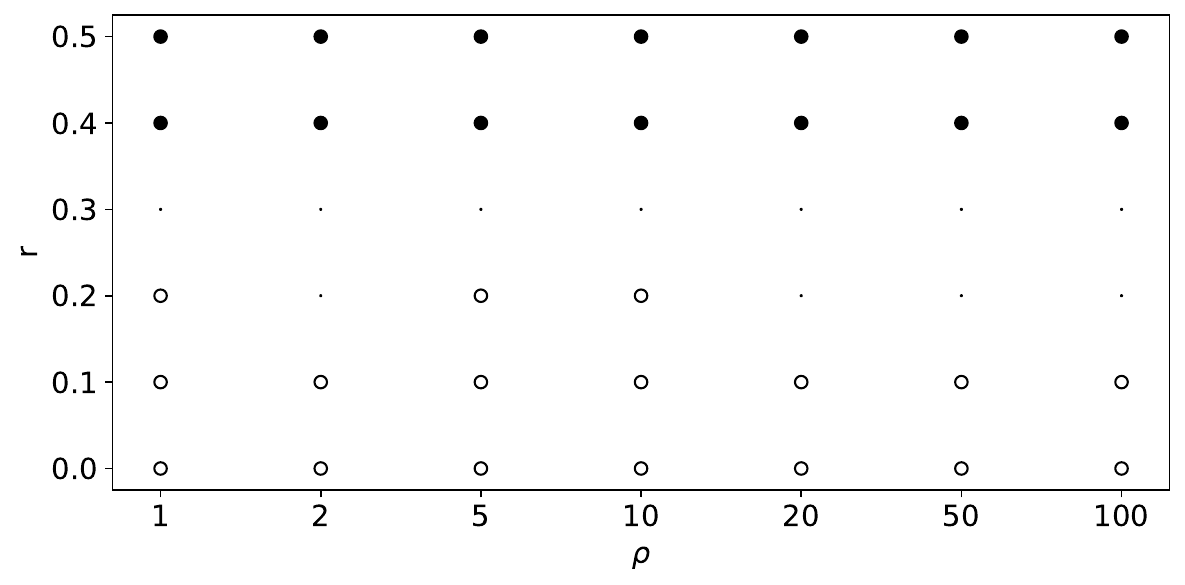}
        \caption{$K$-PDR vs $K$-KMM, $p=0.25$}
    \end{subfigure}
    &
    \begin{subfigure}[b]{0.3\linewidth}
         \includegraphics[width=\textwidth]{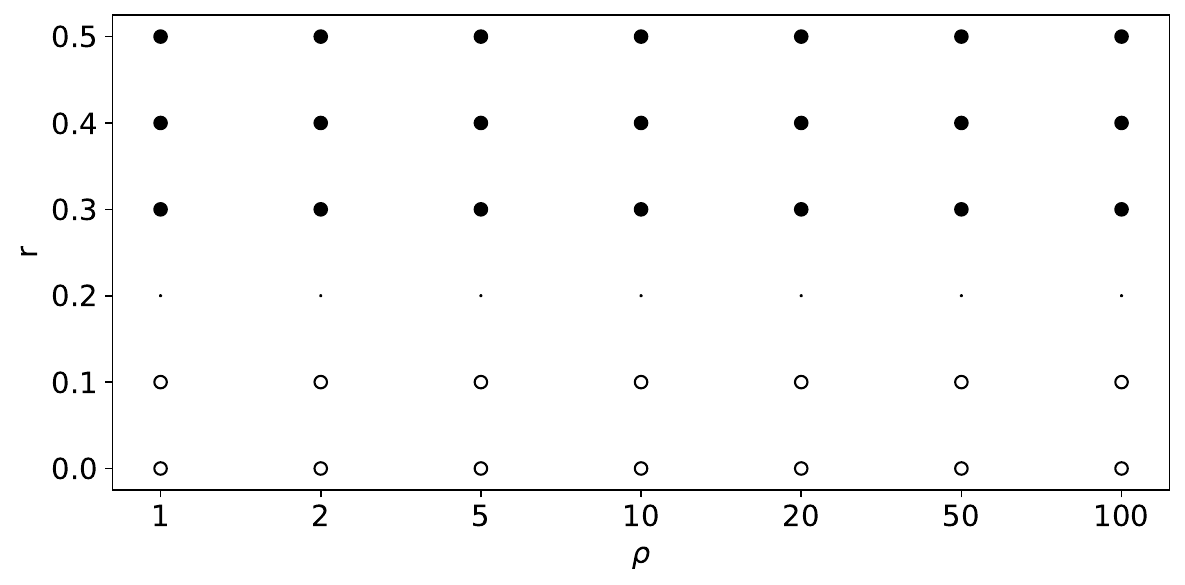}
        \caption{$K$-PDR vs $K$-KMM, $p=0.5$}
    \end{subfigure}
    &
    \begin{subfigure}[b]{0.3\linewidth}
         \includegraphics[width=\textwidth]{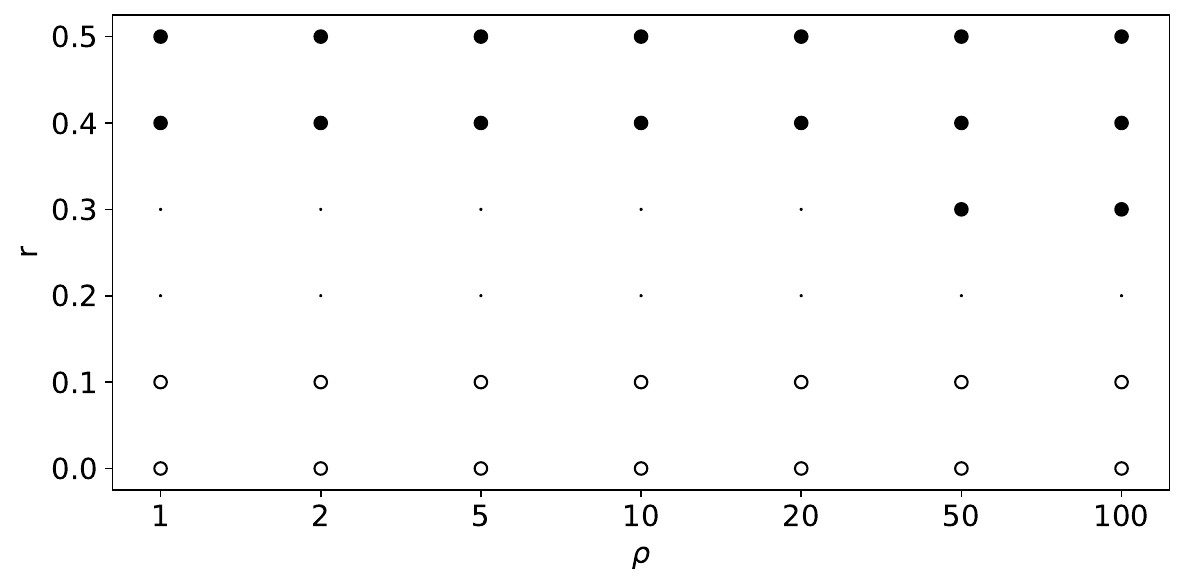}
        \caption{$K$-PDR vs $K$-KMM, $p=0.75$}
    \end{subfigure}
    \\
    \begin{subfigure}[b]{0.3\linewidth}
         \includegraphics[width=\textwidth]{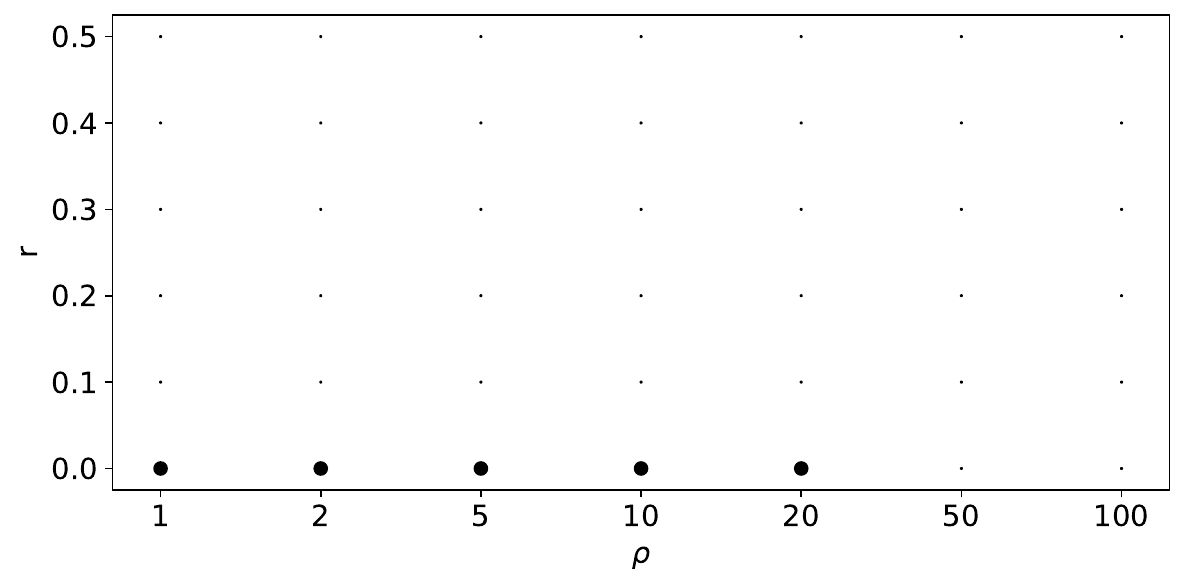}
        \caption{IRBL2 vs $K$-PDR, $p=0.25$}
    \end{subfigure}
    &
    \begin{subfigure}[b]{0.3\linewidth}
         \includegraphics[width=\textwidth]{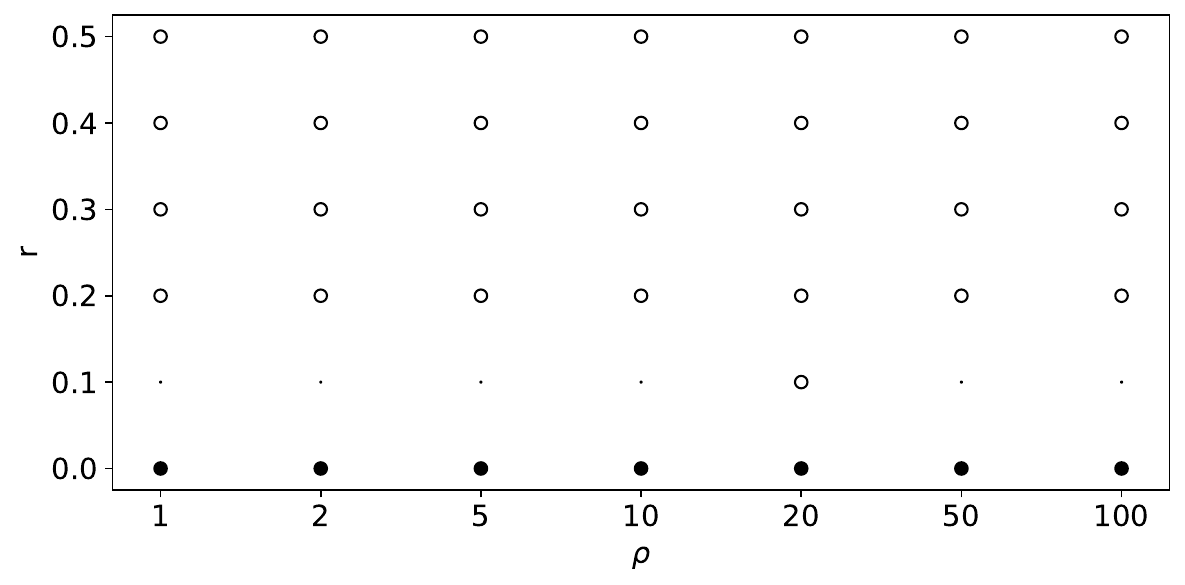}
        \caption{IRBL2 vs $K$-PDR, $p=0.5$}
    \end{subfigure}
    &
    \begin{subfigure}[b]{0.3\linewidth}
         \includegraphics[width=\textwidth]{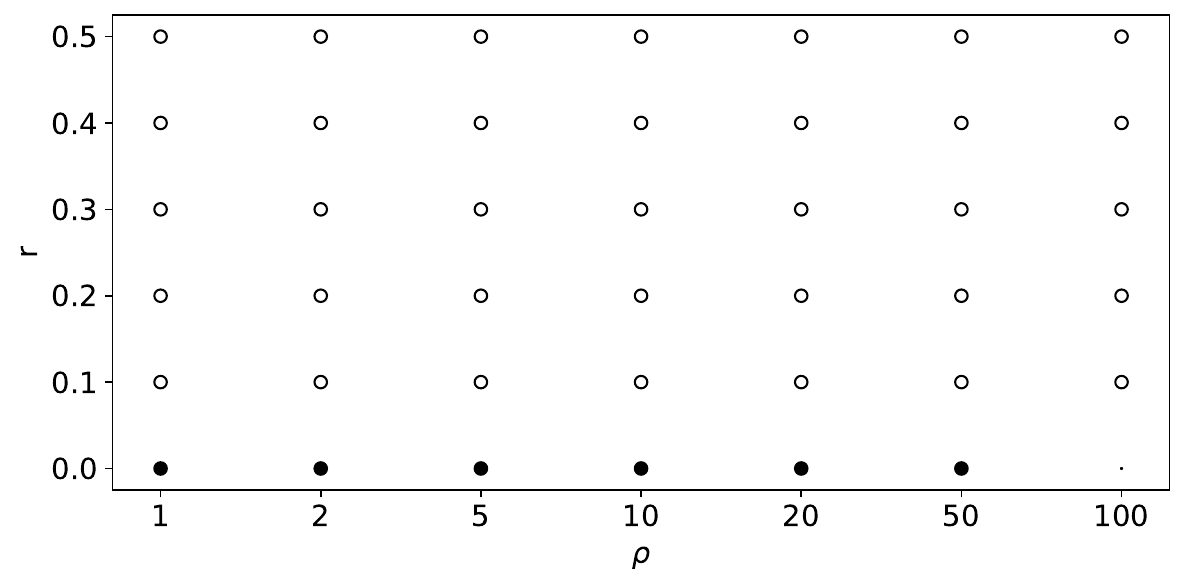}
        \caption{IRBL2 vs $K$-PDR, $p=0.75$}
    \end{subfigure}
    \\
\end{tabular}
\captionsetup{justification=justified}
\caption{Results of the Wilcoxon signed rank test computed on all datasets. Each figure compares one competitor versus another for a given trusted data ratio. Figures in the same row are the same competitors against different cases of trusted data ratio: $p=0.25$, $p=0.5$, $p=0.75$. In each figure ``$\circ$'', ``$\cdot$'' and ``$\bullet$'' indicate respectively a win, a tie, or a loss of the first competitor compared to the second competitor, the vertical axis is $r$, and the horizontal axis is $\rho$.}
\label{wilcoxons}
\end{figure}

Figure \ref{wilcoxons} presents six graphics, each reporting the Wilcoxon test that evaluates one competitor against another based on the accuracy over all datasets. These graphics form a grid with the horizontal axis representing the label noise strength $r$ and the vertical axis representing the sub-sampling ratio $\rho$. On each point of a grid $(r, \rho)$, a Wilcoxon rank-signed test \cite{wilcoxon1992individual} is conducted on two competitors for all datasets to determine if there is a significant difference in accuracy between them. If the first competitor wins, ``$\circ$'' is placed on the grid at this location, ``$\cdot$'' and ``$\bullet$'' indicate a tie or a loss, respectively.

Figure \ref{wilcoxons} allows a point-wise analysis between different competitors, but most notably, it uses a robust statistical test to draw statistically significant conclusions. IRBL2 is not able to improve on IRBL in all tested cases, as seen by the numerous number of ties. $K$-PDR is a better approach than $K$-KMM as the only times $K$-PDR loses to $K$-KMM, $K$-KMM get also beaten by the Trusted Only baseline (see Figure \ref{additional-wilcoxons} in Appendices). Finally, IRBL2 (or IRBL) is a better approach than $K$-PDR when untrusted data are noisy as soon as there is enough trusted data such that the trusted concept can be learned accurately enough to reweight untrusted samples.

\section{Discussions}
\label{discussions}

The previously presented results can be quite underwhelming for both proposed methods, so we propose to discuss these results in this section.

First, results showed that $K$-PDR could substantially improve on $K$-KMM, proving the advantage of using classifiers instead of kernels, which are more powerful tools as they require fewer assumptions about the data distribution and are fitter for learning from heterogeneous data such as tabular data. It also highlights a typical result in real-life ML applications where non-parametric methods work better than parametric methods.

Secondly, results showed no improvements brought by IRBL2 over IRBL on class-conditional shift. Indeed, IRBL might already be able to take into account the uncertainty about $\mathbb{P}(X)$ provided by PDR in the predicted probabilities $\mathbb{P}(Y \mid X)$ thanks to well-calibrated classifiers, nullifying the effect of IRBL2. For example, in unusual regions of the trusted feature space seen in the untrusted dataset, the value of $\mathbb{P}_T(Y \mid X)$ might tend to be around $\frac{1}{K}$ to account the classifier uncertainty and will assign low weights to such samples already without the aid of a specific classifier for that. Meanwhile, $K$-PDR was clearly able to improve from PDR on label noise corruptions.

Furthermore, the proposed experiments for class-conditional shifts might only reproduce some cases of dataset shifts, especially shifts with out-of-distribution data \cite{yang2021generalized}. With the proposed design of synthetic class-conditional shift in our per-class cluster sub-sampling, we did not introduce data points in foreign feature spaces. In our experiments, all points from the untrusted dataset are in-distribution. We only under-represented some sub-populations. This corruption could lead to biased models for individuals from minority groups. However, it is a more manageable situation to deal with than having different groups in the trusted and untrusted dataset. Moreover, the chosen evaluation metric, Cohen's kappa $\kappa$, might not be able to detect such biases, as this metric only evaluates predictive performance.

Finally, the calibration efficiency of the HGBT trees in these experiments greatly impacts the efficiency of every tested biquality-learning algorithm. This benchmark has not explored different calibration techniques other than Isotonic Regression.

Obviously, this benchmark is not a definitive answer to the problem as we could have extended the experiments to other corruptions in real-world untrusted datasets, such as data poisoning \cite{steinhardt2017certified} or class imbalance \cite{japkowicz2002class}.

\section{Conclusion}
\label{conclusion}

In this paper, we have shown the capabilities of the biquality learning framework to design algorithms able to handle closed-set distribution shifts by having access to a trusted and untrusted dataset at training time. We proposed two biquality learning algorithms, IRBL2 and $K$-PDR, inspired respectively by the \textit{label noise} and the \textit{covariate shift} literature. We reviewed distribution shift sources and their hierarchy and proposed two novel methods to synthetically create concept drift and class-conditional shifts in real-world datasets. Throughout extensive experiments, we benchmarked many competitors from the state-of-the-art. We opened some discussions on the results and assessed that the development of biquality learning algorithms robust to distributional changes of closed sets, despite the presented results, remains an interesting problem for future research in this area.

\newpage

\section*{Declarations}

\subsection*{Funding}

Pierre Nodet, Vincent Lemaire, Alexis Bondu, and Antoine Cornuéjols received funding from Orange SA.

\subsection*{Competing Interests}

Pierre Nodet, Vincent Lemaire, and Alexis Bondu have received research support from Orange SA.
Antoine Cornuéjols has received research support from AgroParisTech and Orange SA.

\subsection*{Ethics approval}

Not Applicable.

\subsection*{Consent to participate}

All authors have read and approved the final manuscript.

\subsection*{Consent for publication}

Not Applicable.

\subsection*{Availability of data and materials}

All data used in this study are available publicly online on openML: \url{https://www.openml.org/}

\subsection*{Code availability}

Code is available at the following url: \url{https://github.com/pierrenodet/blds}

\subsection*{Authors' contributions}

Pierre Nodet, Vincent Lemaire, Alexis Bondu, and Antoine Cornuéjols contributed to the manuscript equally.

\newpage

\bibliography{references.bib}
\bibliographystyle{splncs04}

\newpage

\appendix

\section{Additional Tables}

\begin{table}[!h]
\caption{Table of conversion between $p$ and the actual ratio of trusted data (actual).}
\footnotesize
\centering
\setlength{\tabcolsep}{2pt}
\begin{tabular}{llr}
\toprule
Dataset & $p$ & actual \\
\midrule
\multirow[t]{3}{*}{ad} & 0.25 & 0.028 \\
 & 0.50 & 0.041 \\
 & 0.75 & 0.112 \\
\multirow[t]{3}{*}{artificial-chars} & 0.25 & 0.010 \\
 & 0.50 & 0.048 \\
 & 0.75 & 0.273 \\
\multirow[t]{3}{*}{bank-marketing} & 0.25 & 0.003 \\
 & 0.50 & 0.004 \\
 & 0.75 & 0.032 \\
\multirow[t]{3}{*}{click} & 0.25 & 0.165 \\
 & 0.50 & 0.328 \\
 & 0.75 & 0.045 \\
\multirow[t]{3}{*}{connect-4} & 0.25 & 0.003 \\
 & 0.50 & 0.009 \\
 & 0.75 & 0.041 \\
\multirow[t]{3}{*}{dna} & 0.25 & 0.020 \\
 & 0.50 & 0.032 \\
 & 0.75 & 0.044 \\
\multirow[t]{3}{*}{eeg} & 0.25 & 0.010 \\
 & 0.50 & 0.032 \\
 & 0.75 & 0.093 \\
\multirow[t]{3}{*}{electricity} & 0.25 & 0.003 \\
 & 0.50 & 0.004 \\
 & 0.75 & 0.024 \\
\multirow[t]{3}{*}{first-ord-theo-prov} & 0.25 & 0.015 \\
 & 0.50 & 0.082 \\
 & 0.75 & 0.271 \\
\multirow[t]{3}{*}{gas-drift} & 0.25 & 0.003 \\
 & 0.50 & 0.005 \\
 & 0.75 & 0.013 \\
\multirow[t]{3}{*}{har} & 0.25 & 0.007 \\
 & 0.50 & 0.009 \\
 & 0.75 & 0.015 \\
\multirow[t]{3}{*}{hiva} & 0.25 & 0.015 \\
 & 0.50 & 0.070 \\
 & 0.75 & 0.214 \\
\multirow[t]{3}{*}{ibn-sina} & 0.25 & 0.002 \\
 & 0.50 & 0.003 \\
 & 0.75 & 0.004 \\
\multirow[t]{3}{*}{ijcnn1} & 0.25 & 0.002 \\
 & 0.50 & 0.003 \\
 & 0.75 & 0.008 \\
\multirow[t]{3}{*}{isolet} & 0.25 & 0.015 \\
 & 0.50 & 0.015 \\
 & 0.75 & 0.048 \\
\multirow[t]{3}{*}{japanese-vowels} & 0.25 & 0.007 \\
 & 0.50 & 0.009 \\
 & 0.75 & 0.027 \\
\multirow[t]{3}{*}{jm1} & 0.25 & 0.016 \\
 & 0.50 & 0.016 \\
 & 0.75 & 0.149 \\
\multirow[t]{3}{*}{ldpa} & 0.25 & 0.001 \\
 & 0.50 & 0.004 \\
 & 0.75 & 0.026 \\
\multirow[t]{3}{*}{letter} & 0.25 & 0.007 \\
 & 0.50 & 0.012 \\
 & 0.75 & 0.040 \\
    \bottomrule
\end{tabular}
\setlength{\tabcolsep}{2pt}
\begin{tabular}{llr}
\toprule
Dataset & $p$ & actual \\
\midrule
\multirow[t]{3}{*}{magic-telescope} & 0.25 & 0.003 \\
 & 0.50 & 0.004 \\
 & 0.75 & 0.015 \\
\multirow[t]{3}{*}{mnist} & 0.25 & 0.001 \\
 & 0.50 & 0.002 \\
 & 0.75 & 0.005 \\
\multirow[t]{3}{*}{mozilla4} & 0.25 & 0.003 \\
 & 0.50 & 0.005 \\
 & 0.75 & 0.013 \\
\multirow[t]{3}{*}{mushroom} & 0.25 & 0.006 \\
 & 0.50 & 0.008 \\
 & 0.75 & 0.009 \\
\multirow[t]{3}{*}{musk} & 0.25 & 0.009 \\
 & 0.50 & 0.019 \\
 & 0.75 & 0.120 \\
\multirow[t]{3}{*}{nomao} & 0.25 & 0.002 \\
 & 0.50 & 0.003 \\
 & 0.75 & 0.005 \\
\multirow[t]{3}{*}{pendigits} & 0.25 & 0.007 \\
 & 0.50 & 0.008 \\
 & 0.75 & 0.011 \\
\multirow[t]{3}{*}{phishing} & 0.25 & 0.006 \\
 & 0.50 & 0.008 \\
 & 0.75 & 0.009 \\
\multirow[t]{3}{*}{phoeneme} & 0.25 & 0.013 \\
 & 0.50 & 0.016 \\
 & 0.75 & 0.115 \\
\multirow[t]{3}{*}{poker} & 0.25 & 0.002 \\
 & 0.50 & 0.005 \\
 & 0.75 & 0.013 \\
\multirow[t]{3}{*}{satimage} & 0.25 & 0.007 \\
 & 0.50 & 0.009 \\
 & 0.75 & 0.020 \\
\multirow[t]{3}{*}{skin-seg} & 0.25 & 0.000 \\
 & 0.50 & 0.001 \\
 & 0.75 & 0.001 \\
\multirow[t]{3}{*}{spam} & 0.25 & 0.013 \\
 & 0.50 & 0.015 \\
 & 0.75 & 0.033 \\
\multirow[t]{3}{*}{splice} & 0.25 & 0.017 \\
 & 0.50 & 0.019 \\
 & 0.75 & 0.034 \\
\multirow[t]{3}{*}{svmguide3} & 0.25 & 0.160 \\
 & 0.50 & 0.364 \\
 & 0.75 & 0.607 \\
\multirow[t]{3}{*}{usps} & 0.25 & 0.007 \\
 & 0.50 & 0.009 \\
 & 0.75 & 0.014 \\
\multirow[t]{3}{*}{web} & 0.25 & 0.002 \\
 & 0.50 & 0.004 \\
 & 0.75 & 0.009 \\
\multirow[t]{3}{*}{zebra} & 0.25 & 0.033 \\
 & 0.50 & 0.122 \\
 & 0.75 & 0.272 \\
 \\
 \\
 \\
\bottomrule
\end{tabular}
\label{table-conversion-trusted}
\end{table}

\begin{table}[!h]
\caption{Table of conversion between $\rho$ and the actual ratio between the number of samples after and before subsampling (actual).}
\tiny
\centering
\setlength{\tabcolsep}{2pt}
\begin{tabular}{llr}
\toprule
Dataset & $\rho$ & actual \\
\midrule
\multirow[t]{7}{*}{ad} & 1 & 1.000 \\
 & 2 & 0.922 \\
 & 5 & 0.877 \\
 & 10 & 0.861 \\
 & 20 & 0.853 \\
 & 50 & 0.849 \\
 & 100 & 0.848 \\
\multirow[t]{7}{*}{artificial-chars} & 1 & 1.000 \\
 & 2 & 0.814 \\
 & 5 & 0.703 \\
 & 10 & 0.666 \\
 & 20 & 0.647 \\
 & 50 & 0.636 \\
 & 100 & 0.632 \\
\multirow[t]{7}{*}{bank-marketing} & 1 & 1.000 \\
 & 2 & 0.896 \\
 & 5 & 0.834 \\
 & 10 & 0.814 \\
 & 20 & 0.803 \\
 & 50 & 0.797 \\
 & 100 & 0.795 \\
\multirow[t]{7}{*}{click} & 1 & 1.000 \\
 & 2 & 0.889 \\
 & 5 & 0.823 \\
 & 10 & 0.800 \\
 & 20 & 0.789 \\
 & 50 & 0.783 \\
 & 100 & 0.780 \\
\multirow[t]{7}{*}{connect-4} & 1 & 1.000 \\
 & 2 & 0.778 \\
 & 5 & 0.645 \\
 & 10 & 0.601 \\
 & 20 & 0.578 \\
 & 50 & 0.565 \\
 & 100 & 0.561 \\
\multirow[t]{7}{*}{dna} & 1 & 1.000 \\
 & 2 & 0.820 \\
 & 5 & 0.711 \\
 & 10 & 0.675 \\
 & 20 & 0.657 \\
 & 50 & 0.646 \\
 & 100 & 0.643 \\
\multirow[t]{7}{*}{eeg} & 1 & 1.000 \\
 & 2 & 0.913 \\
 & 5 & 0.860 \\
 & 10 & 0.843 \\
 & 20 & 0.834 \\
 & 50 & 0.829 \\
 & 100 & 0.827 \\
\multirow[t]{7}{*}{electricity} & 1 & 1.000 \\
 & 2 & 0.775 \\
 & 5 & 0.641 \\
 & 10 & 0.596 \\
 & 20 & 0.573 \\
 & 50 & 0.560 \\
 & 100 & 0.555 \\
\multirow[t]{7}{*}{first-ord-theo-prov} & 1 & 1.000 \\
 & 2 & 0.922 \\
 & 5 & 0.875 \\
 & 10 & 0.859 \\
 & 20 & 0.852 \\
 & 50 & 0.849 \\
 & 100 & 0.847 \\
\multirow[t]{7}{*}{gas-drift} & 1 & 1.000 \\
 & 2 & 0.876 \\
 & 5 & 0.801 \\
 & 10 & 0.777 \\
 & 20 & 0.764 \\
 & 50 & 0.757 \\
 & 100 & 0.754 \\
\multirow[t]{7}{*}{har} & 1 & 1.000 \\
 & 2 & 0.894 \\
 & 5 & 0.830 \\
 & 10 & 0.809 \\
 & 20 & 0.798 \\
 & 50 & 0.792 \\
 & 100 & 0.790 \\
\multirow[t]{7}{*}{hiva} & 1 & 1.000 \\
 & 2 & 0.940 \\
 & 5 & 0.905 \\
 & 10 & 0.893 \\
 & 20 & 0.887 \\
 & 50 & 0.883 \\
 & 100 & 0.882 \\
\multirow[t]{7}{*}{ibn-sina} & 1 & 1.000 \\
 & 2 & 0.929 \\
 & 5 & 0.886 \\
 & 10 & 0.872 \\
 & 20 & 0.865 \\
 & 50 & 0.861 \\
 & 100 & 0.859 \\
  \bottomrule
\end{tabular}
\setlength{\tabcolsep}{2pt}
\begin{tabular}{llr}
\toprule
Dataset & $\rho$ & actual \\
\midrule
\multirow[t]{7}{*}{ijcnn1} & 1 & 1.000 \\
 & 2 & 0.801 \\
 & 5 & 0.682 \\
 & 10 & 0.642 \\
 & 20 & 0.623 \\
 & 50 & 0.611 \\
 & 100 & 0.607 \\
\multirow[t]{7}{*}{isolet} & 1 & 1.000 \\
 & 2 & 0.828 \\
 & 5 & 0.726 \\
 & 10 & 0.691 \\
 & 20 & 0.674 \\
 & 50 & 0.663 \\
 & 100 & 0.662 \\
\multirow[t]{7}{*}{japanese-vowels} & 1 & 1.000 \\
 & 2 & 0.791 \\
 & 5 & 0.666 \\
 & 10 & 0.624 \\
 & 20 & 0.604 \\
 & 50 & 0.591 \\
 & 100 & 0.587 \\
\multirow[t]{7}{*}{jm1} & 1 & 1.000 \\
 & 2 & 0.960 \\
 & 5 & 0.936 \\
 & 10 & 0.928 \\
 & 20 & 0.924 \\
 & 50 & 0.921 \\
 & 100 & 0.920 \\
\multirow[t]{7}{*}{ldpa} & 1 & 1.000 \\
 & 2 & 0.791 \\
 & 5 & 0.665 \\
 & 10 & 0.624 \\
 & 20 & 0.603 \\
 & 50 & 0.590 \\
 & 100 & 0.586 \\
\multirow[t]{7}{*}{letter} & 1 & 1.000 \\
 & 2 & 0.806 \\
 & 5 & 0.690 \\
 & 10 & 0.651 \\
 & 20 & 0.632 \\
 & 50 & 0.620 \\
 & 100 & 0.617 \\
\multirow[t]{7}{*}{magic-telescope} & 1 & 1.000 \\
 & 2 & 0.832 \\
 & 5 & 0.732 \\
 & 10 & 0.698 \\
 & 20 & 0.681 \\
 & 50 & 0.671 \\
 & 100 & 0.668 \\
\multirow[t]{7}{*}{mnist} & 1 & 1.000 \\
 & 2 & 0.861 \\
 & 5 & 0.777 \\
 & 10 & 0.750 \\
 & 20 & 0.736 \\
 & 50 & 0.727 \\
 & 100 & 0.725 \\
\multirow[t]{7}{*}{mozilla4} & 1 & 1.000 \\
 & 2 & 0.861 \\
 & 5 & 0.777 \\
 & 10 & 0.749 \\
 & 20 & 0.735 \\
 & 50 & 0.727 \\
 & 100 & 0.724 \\
\multirow[t]{7}{*}{mushroom} & 1 & 1.000 \\
 & 2 & 0.852 \\
 & 5 & 0.763 \\
 & 10 & 0.734 \\
 & 20 & 0.719 \\
 & 50 & 0.710 \\
 & 100 & 0.707 \\
\multirow[t]{7}{*}{musk} & 1 & 1.000 \\
 & 2 & 0.814 \\
 & 5 & 0.703 \\
 & 10 & 0.666 \\
 & 20 & 0.647 \\
 & 50 & 0.636 \\
 & 100 & 0.632 \\

\multirow[t]{7}{*}{nomao} & 1 & 1.000 \\
 & 2 & 0.931 \\
 & 5 & 0.890 \\
 & 10 & 0.877 \\
 & 20 & 0.870 \\
 & 50 & 0.866 \\
 & 100 & 0.864 \\
\multirow[t]{7}{*}{pendigits} & 1 & 1.000 \\
 & 2 & 0.830 \\
 & 5 & 0.728 \\
 & 10 & 0.694 \\
 & 20 & 0.677 \\
 & 50 & 0.667 \\
 & 100 & 0.664 \\
   \bottomrule
\end{tabular}
\setlength{\tabcolsep}{2pt}
\begin{tabular}{llr}
\toprule
Dataset & $\rho$ & actual \\
\midrule
\multirow[t]{7}{*}{phishing} & 1 & 1.000 \\
 & 2 & 0.851 \\
 & 5 & 0.761 \\
 & 10 & 0.731 \\
 & 20 & 0.716 \\
 & 50 & 0.708 \\
 & 100 & 0.704 \\
\multirow[t]{7}{*}{phoeneme} & 1 & 1.000 \\
 & 2 & 0.802 \\
 & 5 & 0.683 \\
 & 10 & 0.643 \\
 & 20 & 0.623 \\
 & 50 & 0.611 \\
 & 100 & 0.607 \\
\multirow[t]{7}{*}{poker} & 1 & 1.000 \\
 & 2 & 0.762 \\
 & 5 & 0.620 \\
 & 10 & 0.572 \\
 & 20 & 0.549 \\
 & 50 & 0.534 \\
 & 100 & 0.530 \\
\multirow[t]{7}{*}{satimage} & 1 & 1.000 \\
 & 2 & 0.855 \\
 & 5 & 0.768 \\
 & 10 & 0.740 \\
 & 20 & 0.725 \\
 & 50 & 0.717 \\
 & 100 & 0.714 \\
\multirow[t]{7}{*}{skin-seg} & 1 & 1.000 \\
 & 2 & 0.865 \\
 & 5 & 0.784 \\
 & 10 & 0.757 \\
 & 20 & 0.744 \\
 & 50 & 0.736 \\
 & 100 & 0.733 \\
\multirow[t]{7}{*}{spam} & 1 & 1.000 \\
 & 2 & 0.957 \\
 & 5 & 0.931 \\
 & 10 & 0.922 \\
 & 20 & 0.918 \\
 & 50 & 0.915 \\
 & 100 & 0.914 \\
\multirow[t]{7}{*}{splice} & 1 & 1.000 \\
 & 2 & 0.821 \\
 & 5 & 0.713 \\
 & 10 & 0.676 \\
 & 20 & 0.658 \\
 & 50 & 0.648 \\
 & 100 & 0.646 \\
\multirow[t]{7}{*}{svmguide3} & 1 & 1.000 \\
 & 2 & 0.855 \\
 & 5 & 0.769 \\
 & 10 & 0.737 \\
 & 20 & 0.725 \\
 & 50 & 0.718 \\
 & 100 & 0.716 \\
\multirow[t]{7}{*}{usps} & 1 & 1.000 \\
 & 2 & 0.810 \\
 & 5 & 0.697 \\
 & 10 & 0.658 \\
 & 20 & 0.639 \\
 & 50 & 0.628 \\
 & 100 & 0.624 \\
\multirow[t]{7}{*}{web} & 1 & 1.000 \\
 & 2 & 0.802 \\
 & 5 & 0.683 \\
 & 10 & 0.643 \\
 & 20 & 0.623 \\
 & 50 & 0.612 \\
 & 100 & 0.608 \\
\multirow[t]{7}{*}{zebra} & 1 & 1.000 \\
 & 2 & 0.903 \\
 & 5 & 0.846 \\
 & 10 & 0.826 \\
 & 20 & 0.817 \\
 & 50 & 0.811 \\
 & 100 & 0.809 \\
 \\
 \\
 \\
 \\
 \\
 \\
 \\
 \\
 \\
 \\
 \\
 \\
 \\
 \\
 \bottomrule
\end{tabular}
\label{table-conversion-subsampling}
\end{table}

\begin{figure}[!h]
\centering
\captionsetup{justification=centering}
\begin{tabular}{ccc}
    \begin{subfigure}[b]{0.3\linewidth}
         \includegraphics[width=\textwidth]{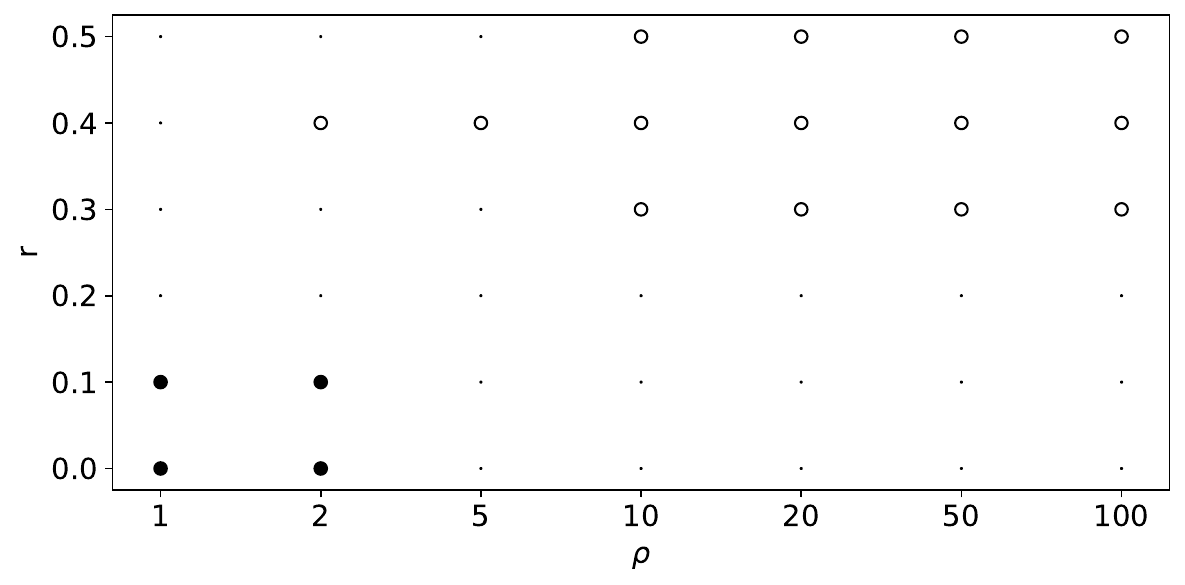}
        \caption{$K$-PDR vs No Correction, $p=0.25$}
    \end{subfigure}
    &
    \begin{subfigure}[b]{0.3\linewidth}
         \includegraphics[width=\textwidth]{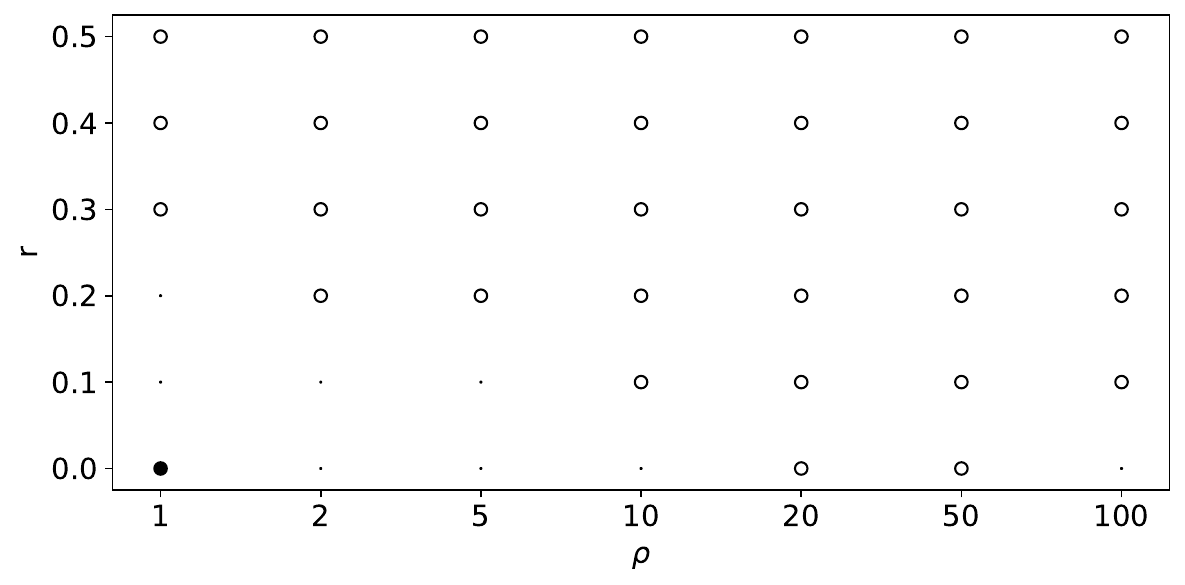}
        \caption{$K$-PDR vs No Correction, $p=0.5$}
    \end{subfigure}
    &
    \begin{subfigure}[b]{0.3\linewidth}
         \includegraphics[width=\textwidth]{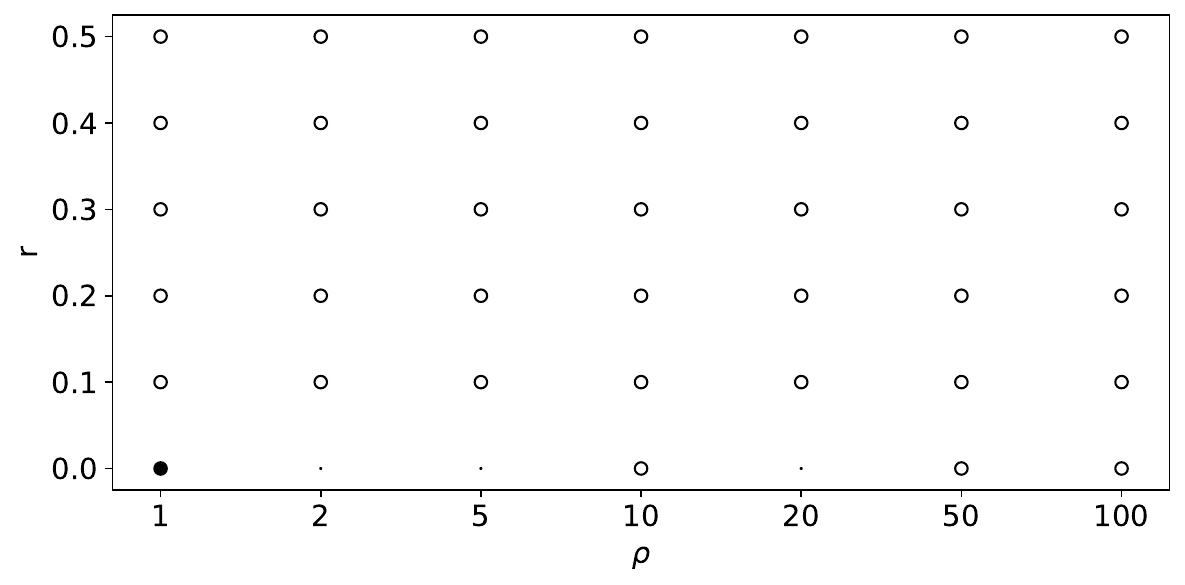}
        \caption{$K$-PDR vs No Correction, $p=0.75$}
    \end{subfigure}
    \\
    \begin{subfigure}[b]{0.3\linewidth}
         \includegraphics[width=\textwidth]{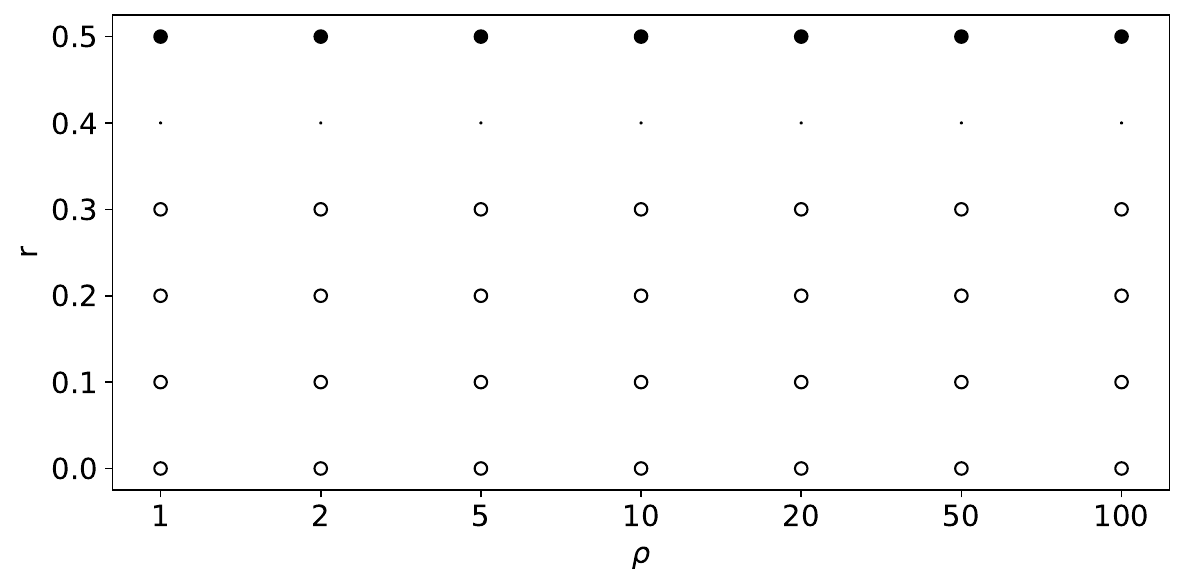}
        \caption{$K$-PDR vs Trusted Only, $p=0.25$}
    \end{subfigure}
    &
    \begin{subfigure}[b]{0.3\linewidth}
         \includegraphics[width=\textwidth]{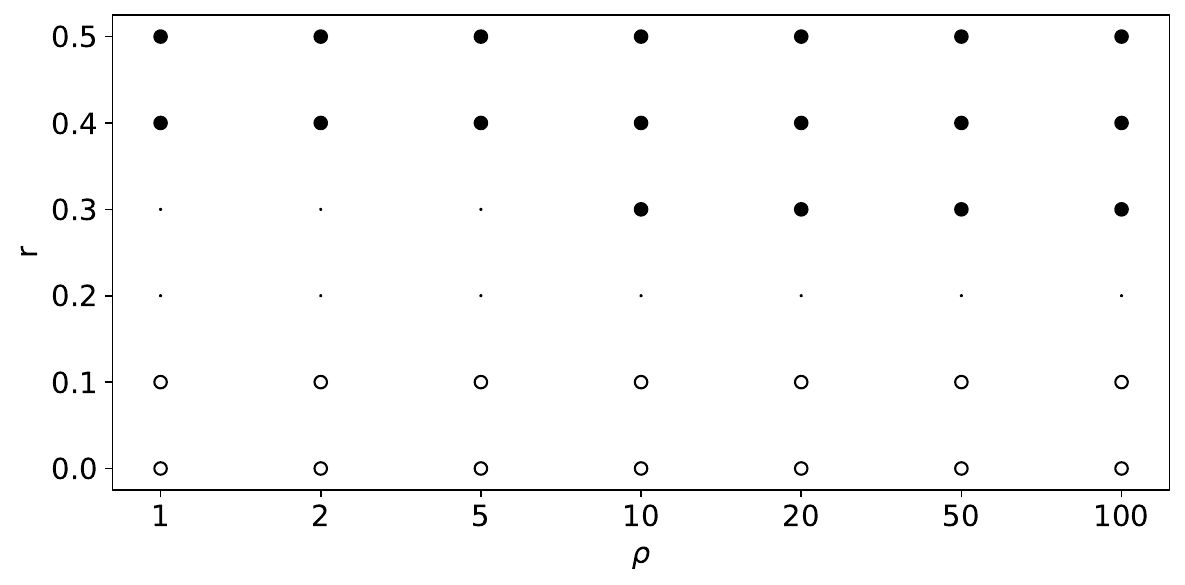}
        \caption{$K$-PDR vs Trusted Only, $p=0.5$}
    \end{subfigure}
    &
    \begin{subfigure}[b]{0.3\linewidth}
         \includegraphics[width=\textwidth]{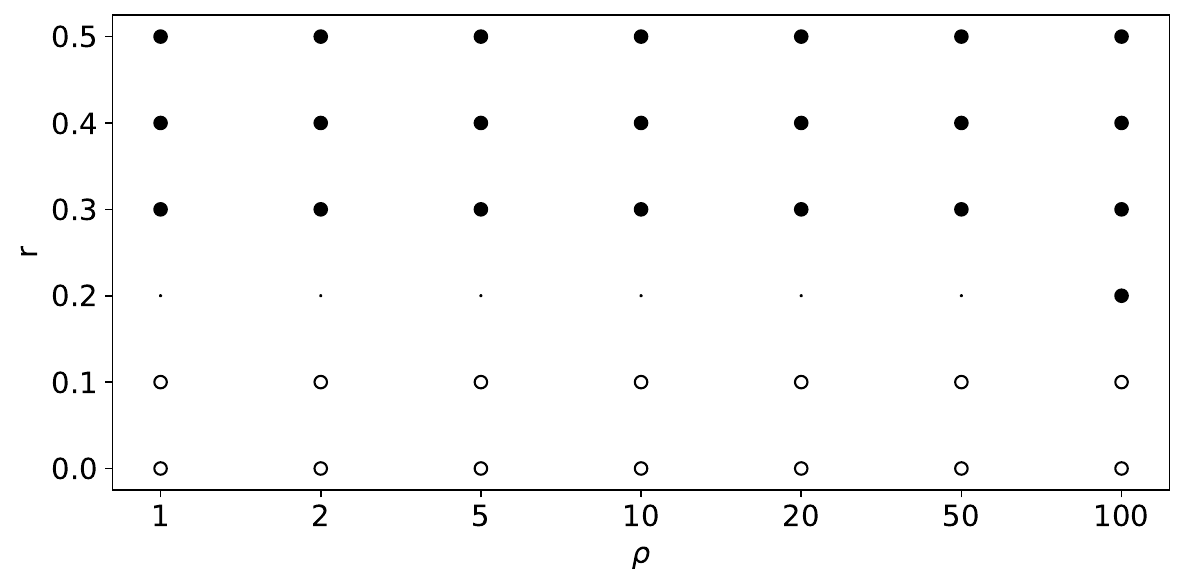}
        \caption{$K$-PDR vs Trusted Only, $p=0.75$}
    \end{subfigure}
    \\
    \begin{subfigure}[b]{0.3\linewidth}
         \includegraphics[width=\textwidth]{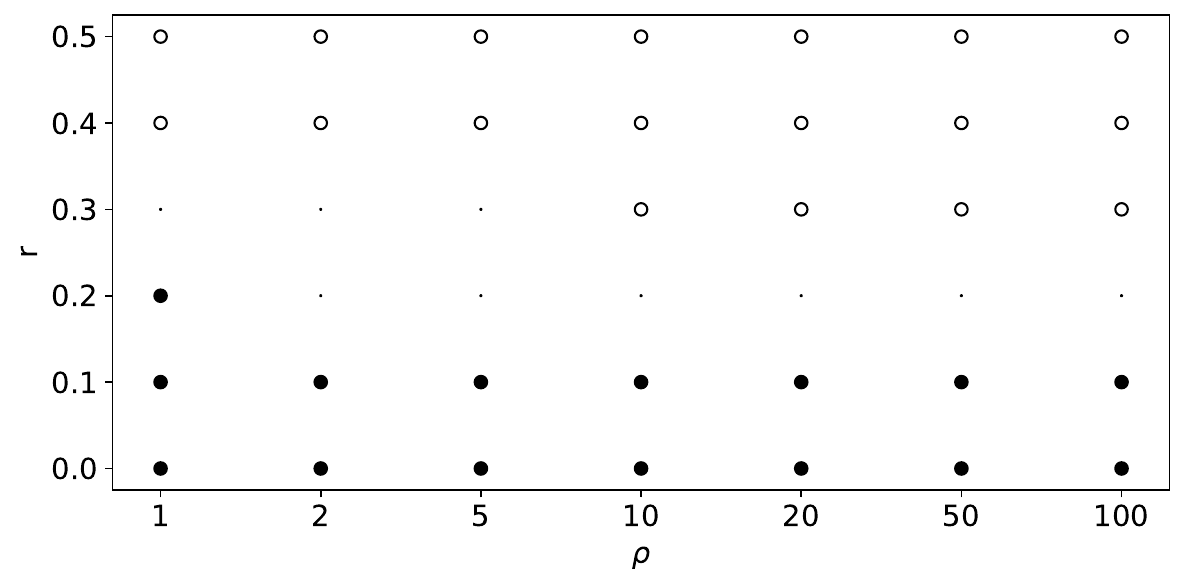}
        \caption{$K$-KMM vs No Correction, $p=0.25$}
    \end{subfigure}
    &
    \begin{subfigure}[b]{0.3\linewidth}
         \includegraphics[width=\textwidth]{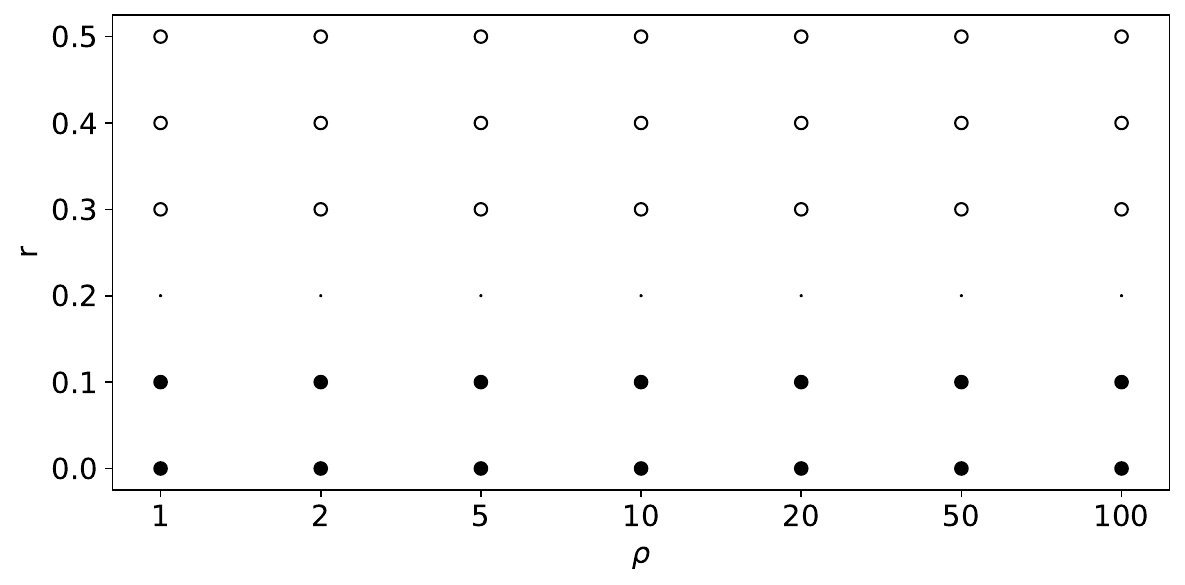}
        \caption{$K$-KMM vs No Correction, $p=0.5$}
    \end{subfigure}
    &
    \begin{subfigure}[b]{0.3\linewidth}
         \includegraphics[width=\textwidth]{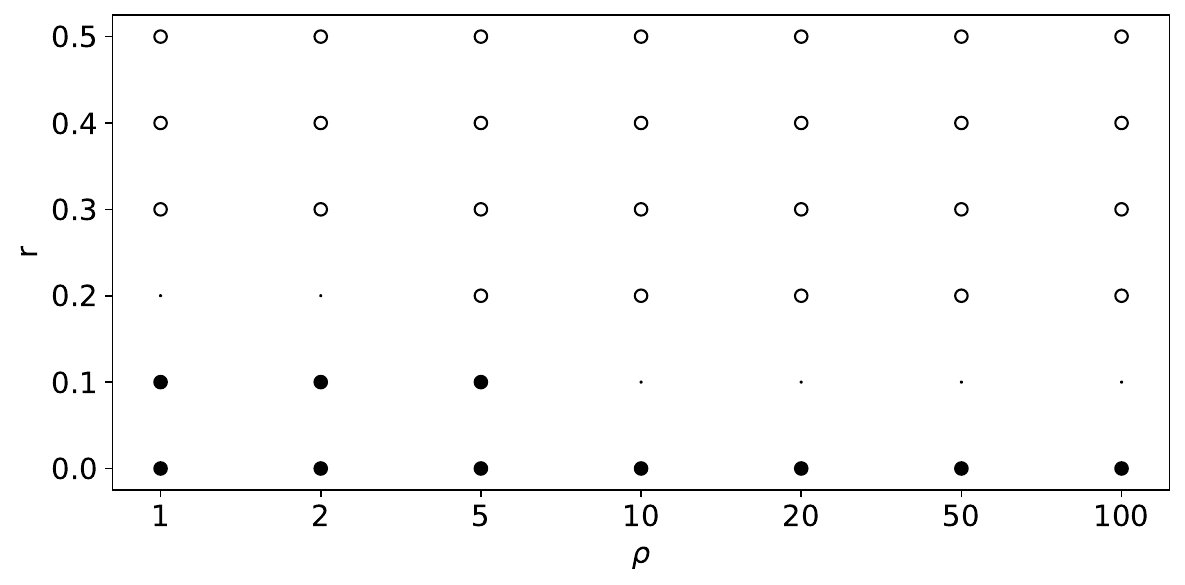}
        \caption{$K$-KMM vs No Correction, $p=0.75$}
    \end{subfigure}
    \\
    \begin{subfigure}[b]{0.3\linewidth}
         \includegraphics[width=\textwidth]{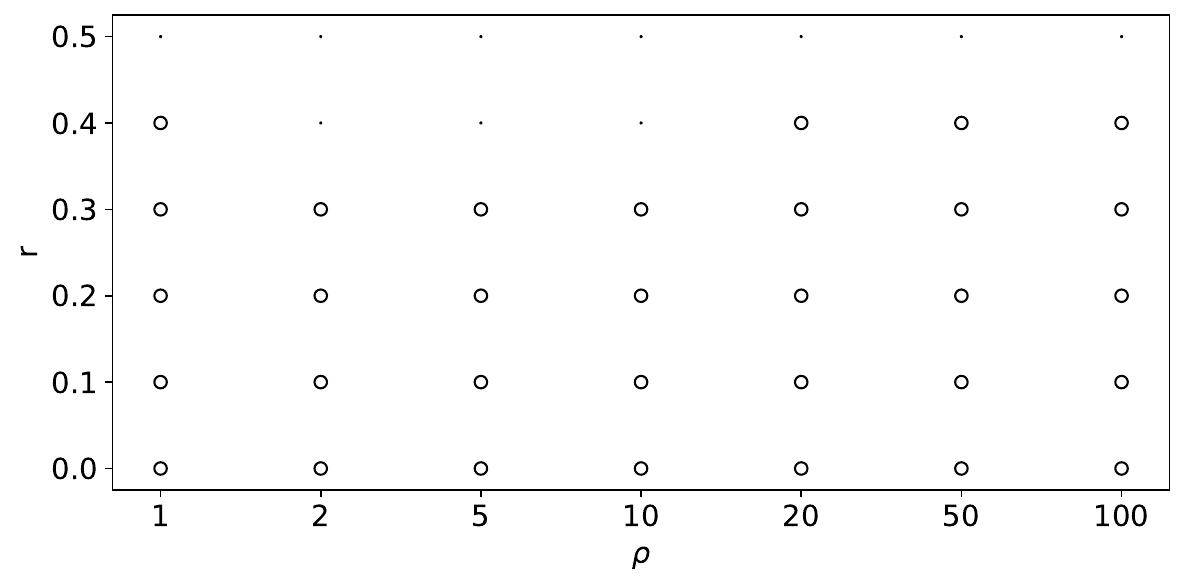}
        \caption{$K$-KMM vs Trusted Only, $p=0.25$}
    \end{subfigure}
    &
    \begin{subfigure}[b]{0.3\linewidth}
         \includegraphics[width=\textwidth]{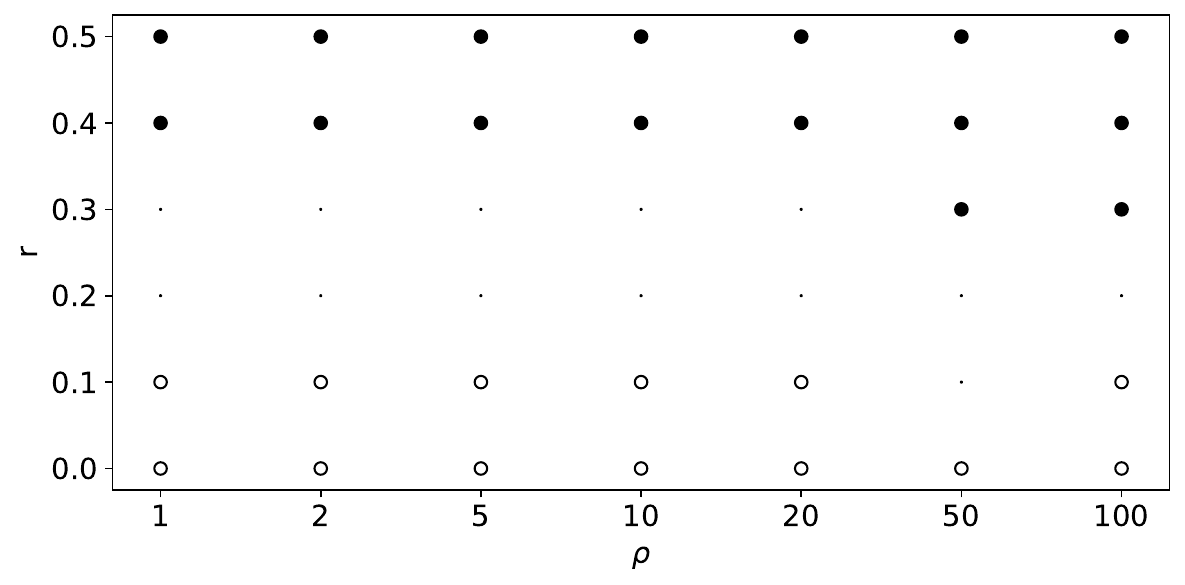}
        \caption{$K$-KMM vs Trusted Only, $p=0.5$}
    \end{subfigure}
    &
    \begin{subfigure}[b]{0.3\linewidth}
         \includegraphics[width=\textwidth]{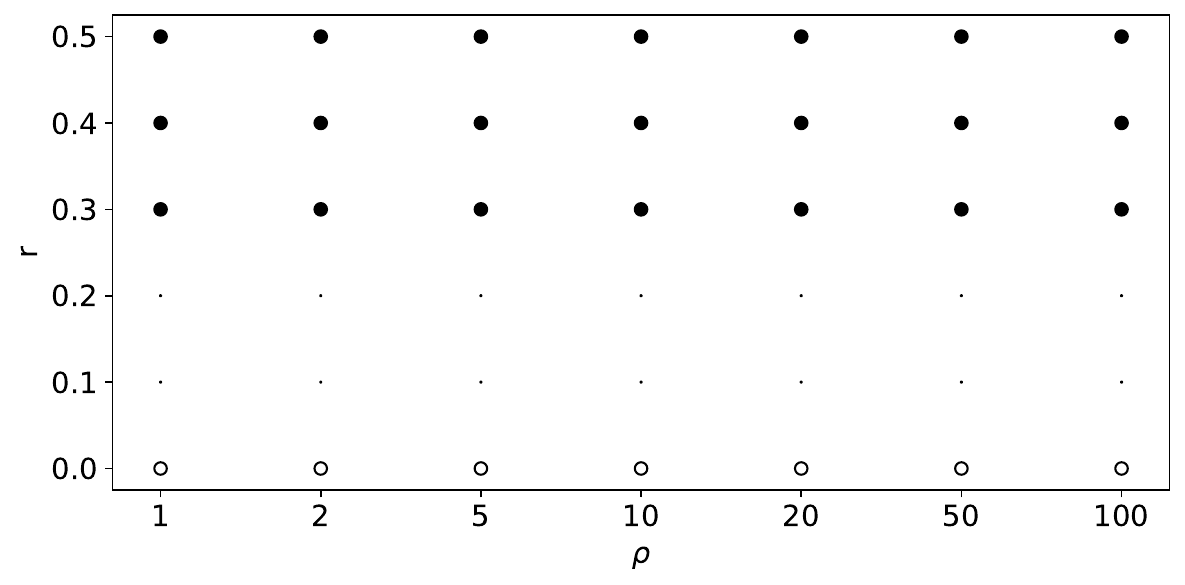}
        \caption{$K$-KMM vs Trusted Only, $p=0.75$}
    \end{subfigure}
    \\
    \begin{subfigure}[b]{0.3\linewidth}
         \includegraphics[width=\textwidth]{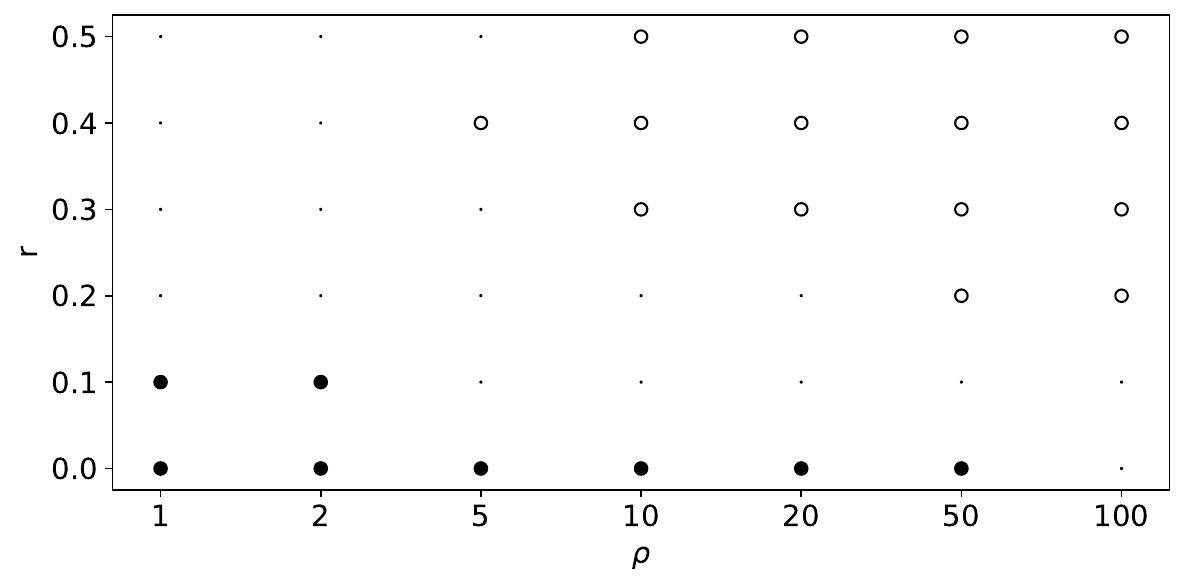}
        \caption{IRBL vs No Correction, $p=0.25$}
    \end{subfigure}
    &
    \begin{subfigure}[b]{0.3\linewidth}
         \includegraphics[width=\textwidth]{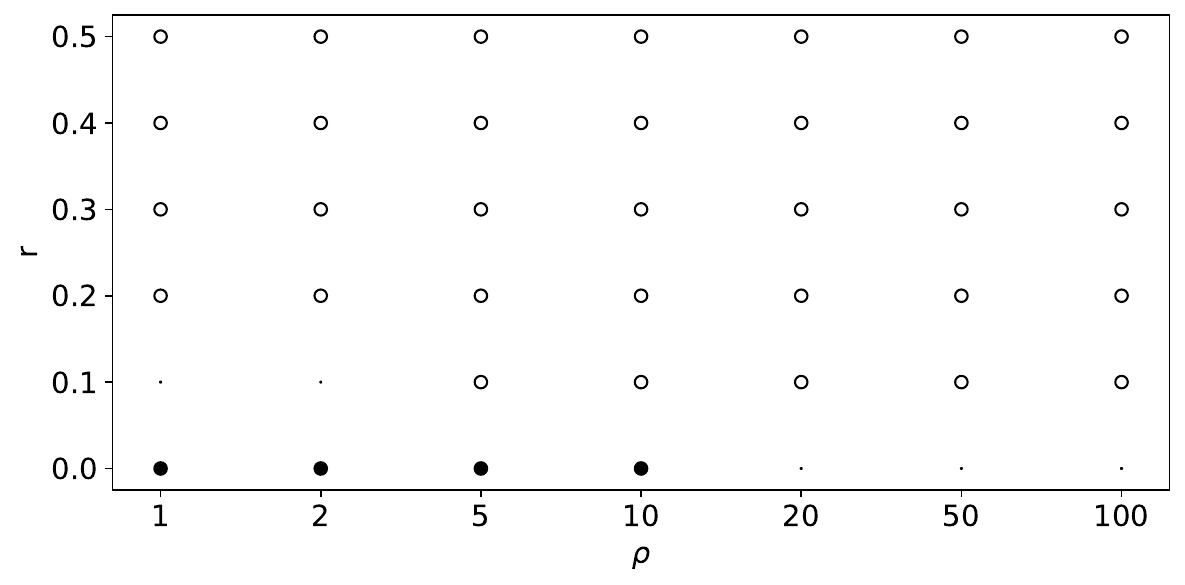}
        \caption{IRBL vs No Correction, $p=0.5$}
    \end{subfigure}
    &
    \begin{subfigure}[b]{0.3\linewidth}
         \includegraphics[width=\textwidth]{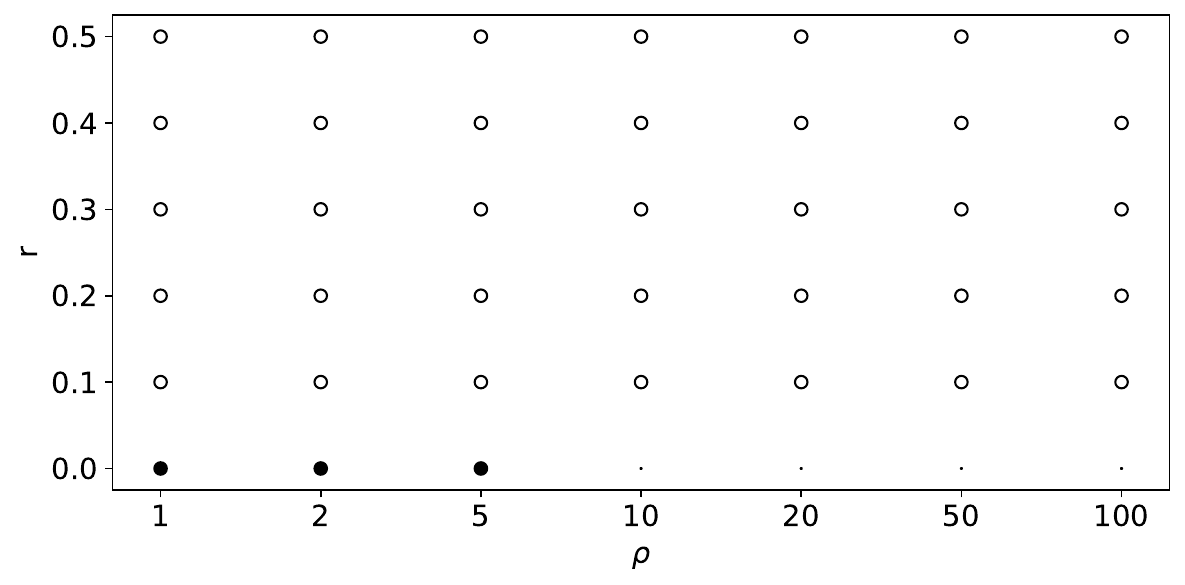}
        \caption{IRBL vs No Correction, $p=0.75$}
    \end{subfigure}
    \\
    \begin{subfigure}[b]{0.3\linewidth}
         \includegraphics[width=\textwidth]{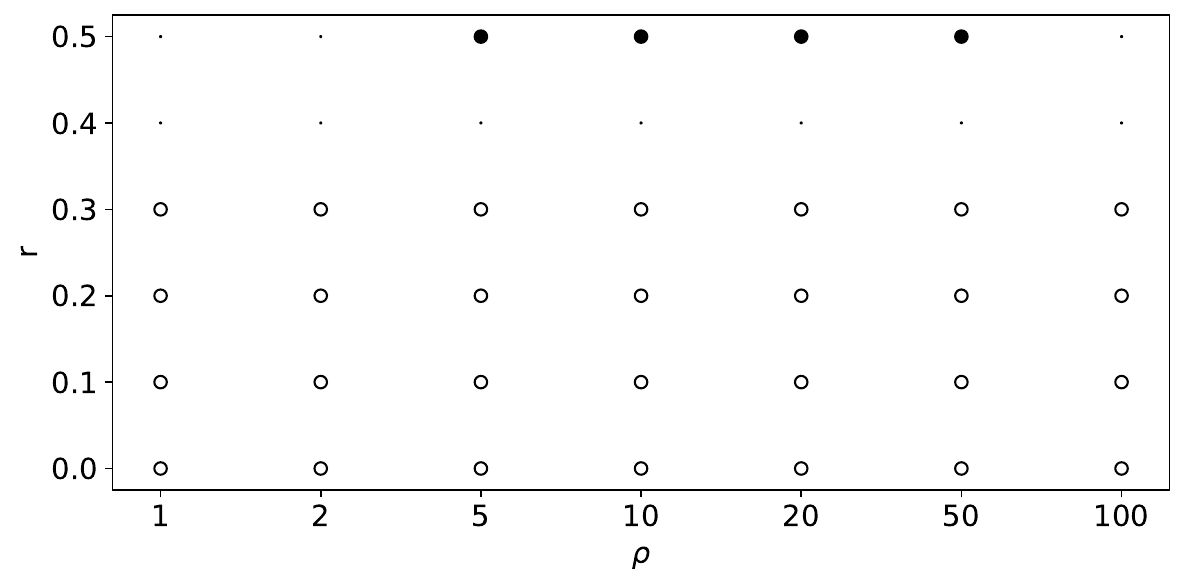}
        \caption{IRBL vs Trusted Only, $p=0.25$}
    \end{subfigure}
    &
    \begin{subfigure}[b]{0.3\linewidth}
         \includegraphics[width=\textwidth]{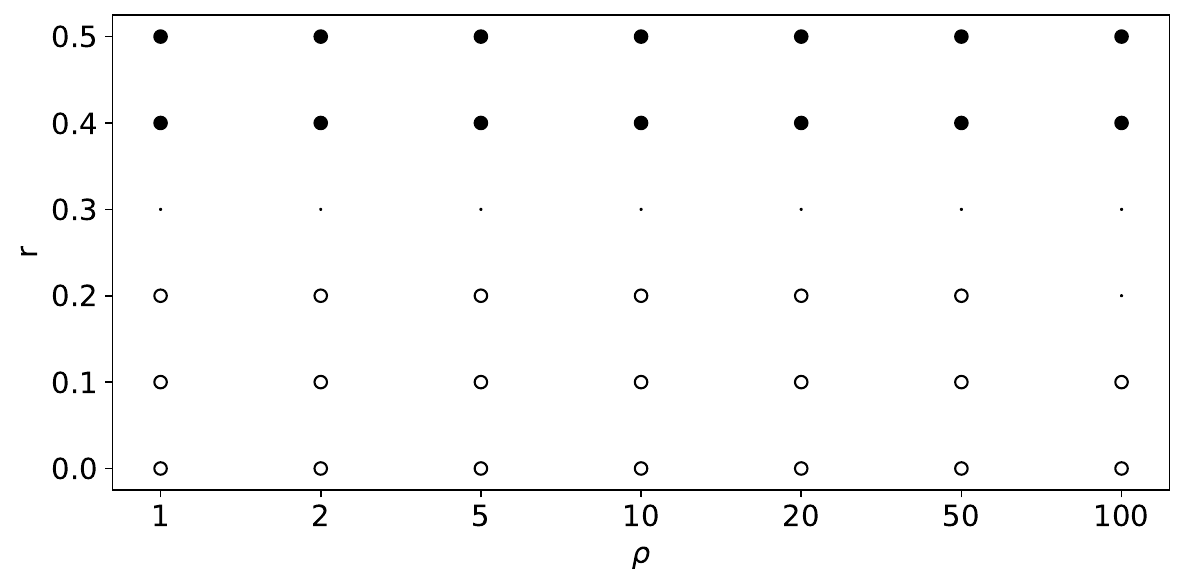}
        \caption{IRBL vs Trusted Only, $p=0.5$}
    \end{subfigure}
    &
    \begin{subfigure}[b]{0.3\linewidth}
         \includegraphics[width=\textwidth]{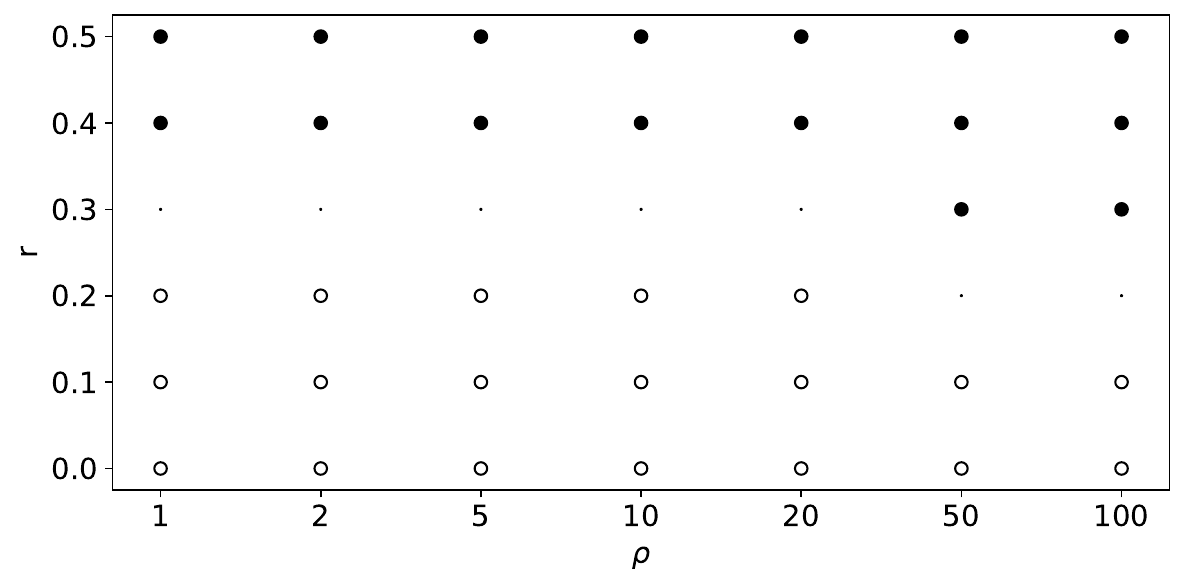}
        \caption{IRBL vs Trusted Only, $p=0.75$}
    \end{subfigure}
    \\
\end{tabular}
\captionsetup{justification=justified}
\caption{Additional results of the Wilcoxon signed rank test computed on all datasets. Each figure compares one competitor versus another for a given trusted data ratio. Figures in the same row are the same competitors against different cases of trusted data ratio: $p=0.25$, $p=0.5$, $p=0.75$. In each figure ``$\circ$'', ``$\cdot$'' and ``$\bullet$'' indicate respectively a win, a tie, or a loss of the first competitor compared to the second competitor, the vertical axis is $r$, and the horizontal axis is $\rho$.}
\label{additional-wilcoxons}
\end{figure}

\end{document}